\documentclass{article}
\usepackage{PRIMEarxiv}
\usepackage[utf8]{inputenc} 
\usepackage[T1]{fontenc}    
\usepackage{url}            
\usepackage{amsfonts}       
\usepackage{nicefrac}       
\usepackage{microtype}      
\usepackage{lipsum}
\usepackage{fancyhdr}       
\usepackage{graphicx}       
\usepackage{algorithm}
\usepackage[marginal]{footmisc}
\usepackage{lmodern}
\usepackage{soul}
\usepackage{MnSymbol}
\usepackage{array}
\usepackage[table,xcdraw]{xcolor}
\usepackage{threeparttable}
\usepackage{booktabs}
\usepackage{multirow}
\usepackage{marvosym}
\usepackage[linewidth = 1pt]{mdframed}
\usepackage[colorlinks, linkcolor=blue, anchorcolor=red,citecolor=blue]{hyperref}

\graphicspath{{figure/}}     

\title{ECRECer: Enzyme Commission Number Recommendation and Benchmarking based on Multiagent Dual-core Learning
\thanks{\textit{\underline{Citation}}: 
\textbf{Authors. Title. Pages.... DOI:000000/11111.}} 
}

\author{
  Zhenkun Shi, Qianqian Yuan, Ruoyu Wang, Hoaran Li, Xiaoping Liao\textsuperscript{\Letter}, Hongwu Ma\textsuperscript{\Letter} \\
  Biodesign Center, Key Laboratory of Systems Microbial Technology \\
  Tianjin Institute of Industrial Biotechnology, Chinese Academy of Sciences, 300308, Tianjin, China\\
  National Technology Innovation Center of Synthetic Biology, 300308, Tianjin, China\\
  \texttt{\{Zhenkun.Shi, yuan\_qq, wangry, lihr, liao\_xp, ma\_hw\}@tib.cas.cn} \\
}

\begin{document}
\maketitle

\begin{abstract}
  Enzyme Commission (EC) numbers, which associate a protein sequence with the biochemical reactions it catalyzes, are essential for the accurate understanding of enzyme functions and cellular metabolism. Many ab-initio computational approaches were proposed to predict EC numbers for given input sequences directly. However, the prediction performance (accuracy, recall, precision), usability, and efficiency of existing methods still have much room to be improved. Here, we report ECRECer, a cloud platform for accurately predicting EC numbers based on novel deep learning techniques. To build ECRECer,  we evaluate different protein representation methods and adopt a protein language model for protein sequence embedding. After embedding, we propose a multi-agent hierarchy deep learning-based framework to learn the proposed tasks in a multi-task manner. Specifically, we used an extreme multi-label classifier to perform the EC prediction and employed a greedy strategy to integrate and fine-tune the final model. Comparative analyses against four representative methods demonstrate that ECRECer delivers the highest performance, which improves accuracy and F1 score by 70\% and 20\% over the state-of-the-the-art, respectively. With ECRECer, we can annotate numerous enzymes in the Swiss-Prot database with incomplete EC numbers to their full fourth level. Take UniPort protein "A0A0U5GJ41" as an example (1.14.-.-), ECRECer annotated it with "1.14.11.38", which supported by further protein structure analysis based on AlphaFold2. Finally, we established a webserver (\href{https://ecrecer.biodesign.ac.cn/}{https://ecrecer.biodesign.ac.cn}) and provided an offline bundle to improve usability.
\end{abstract}


\section{Introduction}

With the widespread adoption of high-throughput methods and high-quality infrastructure in biotechnology and bioindustry, the speed of new protein discovery has increased dramatically. However, this was not followed by a concomitant increase in the speed of protein annotation (see Supplemental, SI Appendix FIGURES, Fig. S2). For example, 5 241 146 sequences were added to TrEMBL in the UniProt database \cite{uniprot2021uniprot} in the single month of March 2021, while only 521 sequences were reviewed and added to Swiss-Prot in the same period \href{./supplementary.pdf}{(see Supplemental, SI Appendix FIGURES, Fig. S2)}. Such a slow speed of protein annotation considerably restricts related research and industrial applications.

Among the multiple and complex protein annotation tasks, one of the crucial steps is enzyme function annotation \cite{ryu2019deep, furnham2009missing}. Annotations of enzyme function provide critical starting points for generating and testing biological hypotheses \cite{furnham2009missing}. Current functional annotations of enzymes describe the biochemistry or process by assigning an Enzyme Commission (EC) number. This is a four-part code associated with a recommended name for the corresponding enzyme-catalyzed reaction that describes the enzyme class, the chemical bond acted on, the reaction, and the substrates \cite{ecmuberwiki2021}. Thus, the primary task of enzyme annotation is to assign an EC number to a given protein sequence. However, as the uncertainty of the assignments for uncharacterized protein sequences is high and biochemical data are relatively sparse, both the speed and the quality of enzyme annotation are considerably restricted.

To achieve improved, rapid and intelligent functional annotation, computational methods were introduced to assign or predict EC numbers. The simplest and most commonly used method is multiple sequence alignment (MSA) \cite{hung2016sequence}, which can yield an appropriate annotation by using similar sequences. Based on this approach, researchers have developed most major EC databases and profile-based methods for the functional annotation of enzymes \cite{yu2009genome, claudel2003enzyme, nursimulu2018improved, zhang2017cofactor}. However, these methods cannot perform annotations for novel proteins with no similar sequences, which is generally the case for newly discovered enzymes. To overcome this restriction, researchers introduced machine learning methods, such as SVM \cite{li2016svm}, KNN \cite{dalkiran2018ecpred}, and hidden Markov model \cite{arakaki2009eficaz} for the functional annotation of enzymes. Although these methods can predict EC numbers even if the given protein sequences have no similar references, the prediction speed and precision are not ideal. Since deep learning has delivered powerful results in many areas \cite{akinosho2020deep,li2020modern, li2021deep, shi2021deep}, more researchers are trying to use deep learning methods to predict EC numbers and significantly improve the precision of functional annotation. However, deep learning methods are prone to overfitting due to an unbalanced distribution of training datasets. In EC number prediction, this leads to prediction results with high precision, medium recall, and low accuracy.

Overall, there has been a steady improvement in computational methods for enzyme annotation \cite{zhang2017cofactor, claudel2003enzyme, shen2007ezypred, ryu2019deep}, but several obstacles still exist that have slowed the progress of computational enzyme function annotation. One of the direct challenges is a lack of publicly available benchmark datasets to evaluate the existing and newly proposed models, which makes it troublesome for the end-user to choose the best method in their production scenario. Another notable challenge is the lack of an efficient and universal protein sequence embedding method. Thus, researchers have to spend large amounts of time on handcrafted feature engineering to encode the sequence, such as functional domain encoding \cite{li2018deepre} and position-specific scoring matrix encoding \cite{An2019}, as encoding quality dramatically impacts the performance of downstream applications \cite{Yang2018}. The third challenge is the lack of an explicitly designed method to deal with this extreme multi-label classification problem (more than 5000 EC numbers in UniProt). Thus, obtaining reliable EC number prediction results is not straightforward, and the prediction performance is not ideal. The fourth noteworthy challenge is the usability of existing tools that need refinement so that the end-user can use them smoothly even with no coding experience.

In this paper, we take a unified approach to address these challenges. For the first challenge, we constructed three standard datasets for benchmarking and evaluation. The datasets contain more than 470,000 distinct labeled protein sequences from Swiss-Prot. To address the second challenge, we introduced the cutting-edge ideology from natural language embedding for protein sequence representation. Firstly, state-of-the-art deep learning methods were evaluated and adopted for universal protein sequence embedding \cite{alley2019unified, rao2020transformer}. Then, we used a feedback mechanism to choose the most suitable method in response to the downstream tasks for optimization. To address the third challenge, we proposed a Dual-core Multiagent Learning Framework (DMLF) for EC number prediction. In DMLF, we formulate the EC number prediction as a three-step hierarchical extreme multi-label classification problem. The first step predicts whether a given protein sequence is an enzyme or not. The second predicts how many functions the enzyme can perform, i.e., multifunctional enzyme prediction. The last step predicts the exact EC number for each enzyme function. We use traditional machine learning methods in the first two steps and a novel deep learning-based extreme multi-label classifier in the last step, then use a greedy strategy to integrate these steps to maximize the EC prediction performance. To address the last challenge, we streamlined the construction process and open-sourced our codes. Moreover, we published a webserver, so that anyone can annotate EC numbers smoothly in high-throughput, whether they have coding experience or not.

\section{Methodology}

This section consists of five subsections. We first formulate the enzyme function annotation problem in the first subsection, and then describe the benchmark data construction process in the second subsection. The third subsection describes our proposed DMLF framework for the benchmark tasks. The fourth subsection describes the baselines. In the last subsection, we describe the evaluation metrics.

\subsection{Problem Formulation}

In order to annotate the enzyme function of a new protein sequence, the initial and basic task is to define whether a given protein is an enzyme. Since there are numerous multifunctional enzymes, the next task to consider is to determine whether the enzyme is monofunctional or multifunctional. If it is multifunctional, the number of functions needs to be classified. After completing the above two tasks, it is necessary is to assign an EC number to each function. Based on these considerations, we proposed three basic tasks for the functional annotation of enzymes, as shown below.

\subsubsection{\textbf{Enzyme or Non-enzyme Annotation.}}
The enzyme or non-enzyme annotation task is formulated as a binary classification problem:

\begin{equation}
    \label{eq:d_task1}
    f: X \rightarrow \{0, 1\}
\end{equation}
where $X=\{x_1, x_2, \cdots, x_n\}, n\geq 1$ represents a group of  protein sequences, and $\{0, 1\}$ is the label indicting whether a given protein is an enzyme .

\subsubsection{\textbf{Multifunctional Enzyme Annotation.}}
Multifunctional enzyme annotation is formulated as a multi-classification problem:

\begin{equation}
    \label{eq:d_task2}
    f: X \rightarrow \{1, 2, \cdots, k \}, 
\end{equation}

where $k$ represents the maximum number of EC number for a given protein. 

\subsubsection{\textbf{Enzyme Commission Number Assignment.}}

The enzyme commission number assignment task is also formulated as a multi-classification problem as defined in Eq. \ref{eq:d_task3}. 

\begin{equation}
    \label{eq:d_task3}
    f: X \rightarrow \{1.1.1.1, 1.1.1.2, 1.1.1.3 , \cdots\}, 
\end{equation}

\subsection{Dataset Description}

To address the first challenge, we constructed three standard datasets \href{./supplementary.pdf}{(Supplemental, SI Appendix Materials and Methods, A. Dataset)}. Similar to previous work [21, 26], these datasets are extracted from the Swiss-Prot database. To simulate real application scenarios as closely as possible, we did not shuffle data randomly. Instead, after data preprocessing \href{./supplementary.pdf}{(See Supplemental, SI Appendix Materials and Methods A. Preprocessing)}, we organized data in chronological order. Specifically, we used a snapshot from Feb 2018 as the training dataset. The training data contains 469,134 distinct sequences in a total 556,825 records, among which 53.56\% are non-enzymes, while the remaining 47.44\% are enzymes. The testing data was extracted from the June 2020 snapshot and sequences that appeared in the training set were excluded. The details are listed in Table \ref{tab:dataset}.

\begin{table}[ht]
  \centering
  \caption{Description of Benchmarking Datasets}
  \label{tab:dataset}
  \begin{tabular}{p{3.5cm}<{\centering}|
                  p{2.0cm}<{\centering}|
                  p{2.0cm}<{\centering}|
                  p{1.6cm}<{\centering}|
                  p{1.7cm}<{\centering}|
                  p{1.8cm}<{\centering}}
      \hline
      \rowcolor[HTML]{EFEFEF} &\multicolumn{2}{c|}{Snapshot}      & \multicolumn{3}{c}{Difference}  \\ \cline{2-6} 
      \rowcolor[HTML]{EFEFEF} 
      \multirow{-2}{*}{ITEM}  & Feb.2018       & Jun.2020   & differ   & added & deleted        \\ \hline
      Records                 & 556,825        & 563,972    & 7147     & -     & -              \\ \hline
      Duplicate Removal       & 469,134        & 476,006    & 6877     & 8033  & 1156           \\ \hline
      Non-enzyme              & 246,567       & 247,324   & 757        & 4454  & 879            \\ \hline
      Enzyme                  & 222,567       & 228,687   & 6120       & 3579  & 277            \\ \hline
      Distinct EC             & 4854          & 5306      & 452        & 644   & 192            \\ \hline
      \end{tabular}
\end{table}

\noindent $\blacksquare$ \textbf{Dataset 1: Enzyme and  Non-enzyme Dataset }

As listed in Table \ref{tab:dataset_enzyme_nonenzyme}, the training set in total has 469,134 records, 222,567 of which are enzymes, and 246,567 are non-enzymes. The testing set contains 7101 records, 3304 of which are enzymes, and the other 3797 are non-enzymes. To make the data more inclusive, we did not filter any sequence in terms of length and homology, which is different from previous studies. An enzyme is labeled as 1 and non-enzyme is labelled as 0. More details about the dataset can be found in the Supplemental, \href{./supplementary.pdf}{SI Appendix Materials and Methods, A. Dataset}.

\begin{table}[htbp]
  \centering
  \caption{Description of Enzyme and Non-enzyme Datasets}
  \label{tab:dataset_enzyme_nonenzyme}
  \begin{tabular}{p{4.7cm}<{\centering}|p{3.7cm}<{\centering}|p{3.7cm}<{\centering}}
  \hline \rowcolor[HTML]{EFEFEF} 
  ITEM       & Training set   & Testing set  \\ \hline
  Enzyme     & 222,567        & 3,304 \\ \hline
  Non-enzyme & 246,567        & 3,797 \\ \hline
  Total      & 469,134        & 7,101 \\ \hline
  \end{tabular}
\end{table}

\noindent $\blacksquare$ \textbf{Dataset 2: Multifunctional Enzyme Dataset }

The multifunctional enzyme dataset only contains enzyme data (225,871 records). The number of EC categories ranges from 1 to 8. The details of the dataset are listed in Table \ref{tab:dataset_quantity}.

\begin{table}[ht]
    \centering
    \caption{Description of the Multifunctional Enzyme Dataset}
    \label{tab:dataset_quantity}
    \begin{tabular}{ccc||ccc}
        \toprule
        \multirow{2}{*}{$\#$EC} & 
        \multicolumn{2}{c||}{Records}& 
        \multirow{2}{*}{$\#$EC} & 
        \multicolumn{2}{c}{Records} \\  
                & Trainset & Testset&     & Trainset & Testset \\ \midrule
        1       & 210788   & 3052   & 5   & 206      & 6      \\
        2       & 9943     & 183    & 6   & 80       & 2      \\ 
        3       & 993      & 53     & 7   & 27       & 1      \\ 
        4       & 525      & 7      & 8   & 5        & 0      \\ 
        
        \bottomrule
    \end{tabular}
\end{table}

\noindent $\blacksquare$ \textbf{Dataset 3: EC number Dataset }

\begin{figure}[htbp]
	\centering	
	\includegraphics[width=0.80\textwidth]{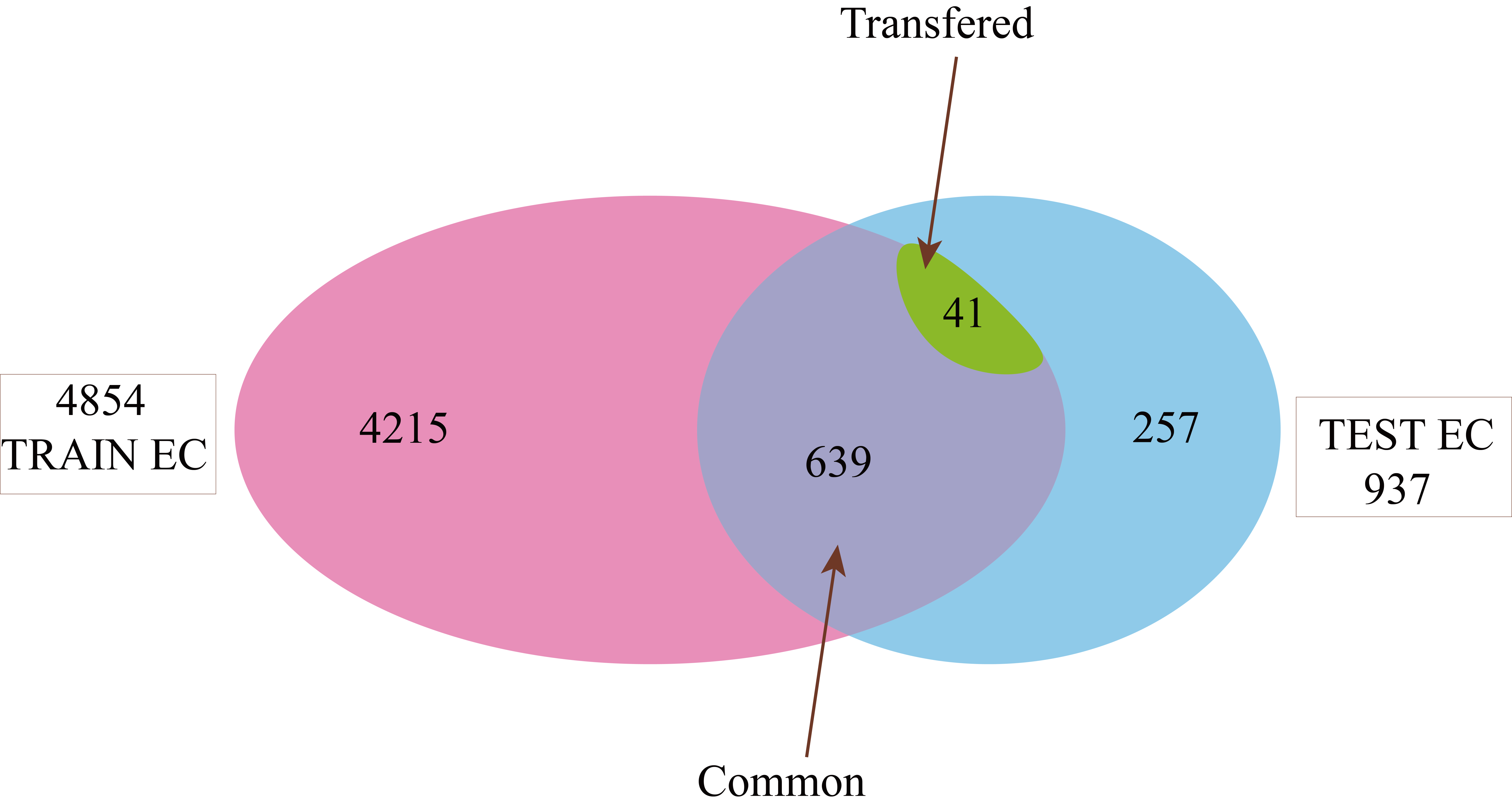}
	\caption{ Venn diagram of the training and testing datasets }
	\label{fig:ec_number}
\end{figure}

Similar to the multifunctional enzyme dataset, the EC number dataset consists of 225,871 enzyme records, 222,567 of which constitute the training dataset, and the remaining 3304 are the testing dataset, covering 5111 EC numbers. As shown in Fig. \ref{fig:ec_number}, the test data include 257 newly added EC numbers compared with the training data, which means that these EC numbers did not appear in the training process, so predictive methods cannot handle this part of the EC numbers. Thus, we excluded the sequences with these 257 EC numbers in the evaluation process. More details about the dataset can be found in \href{./supplementary.pdf}{Supplemental, SI Appendix Materials and Methods, A. Dataset}.

\subsection{Proposed Framework}

\begin{figure*}[htbp]
    \centering	
	\includegraphics[width=0.98\textwidth]{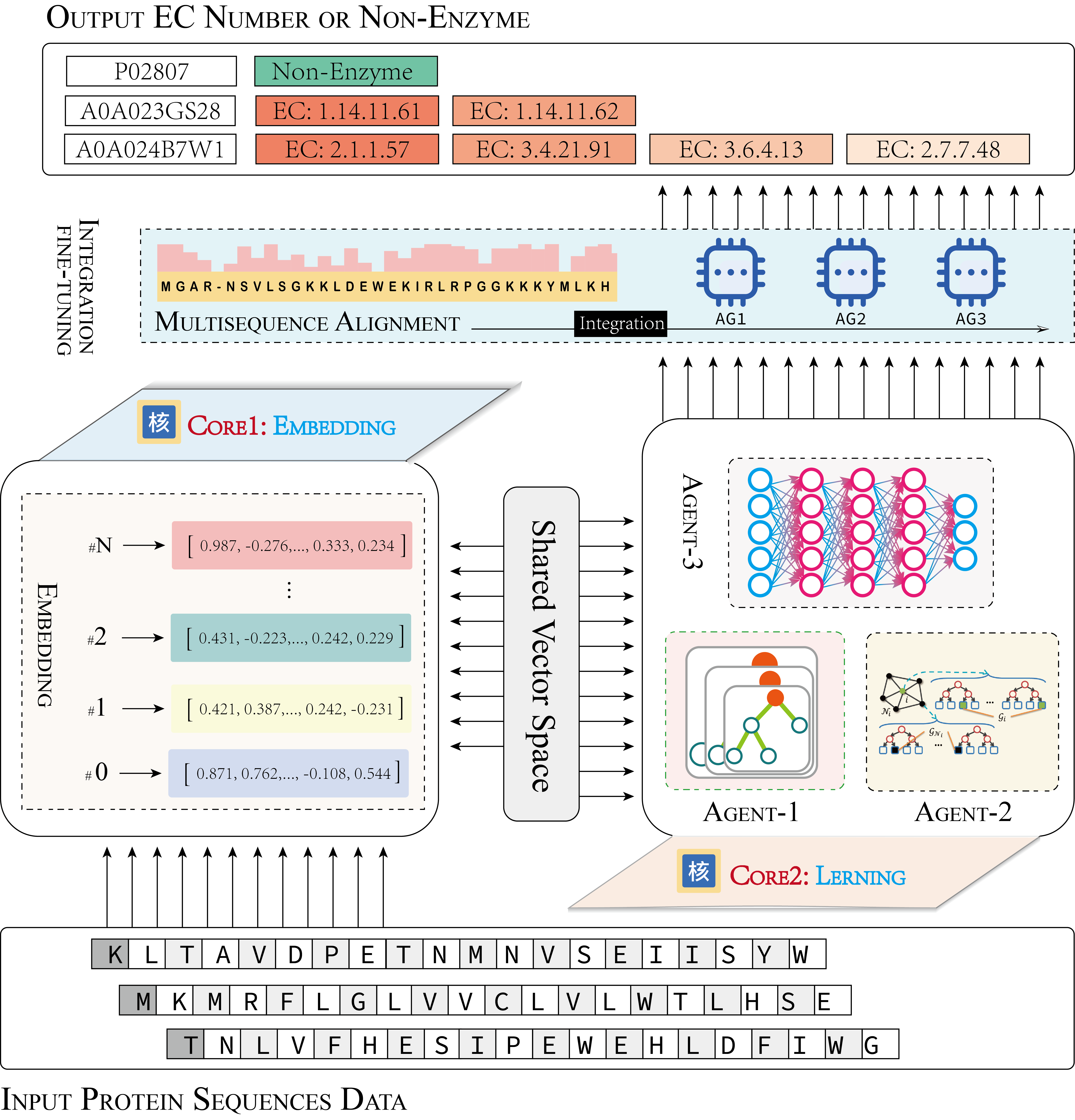}
	\caption{
        DMLF is an explicitly designed dual-core driven framework for EC number prediction. It consists of 2 independent operation units - an embedding core and a learning core. The embedding core is tasked with converting protein sequences into features. The learning core is designed to address the specific biological tasks defined in the problem formulation section. We use different agents to solve different tasks. Agent 1 was designed to solve the enzyme or non-enzyme classification task, agent 2 was designed to solve the multifunctional enzyme prediction task, and agent 3 was designed to solve the EC number assignment task.
	}
	\label{fig:framework}
\end{figure*}

To address the second and third challenges: lack of a generic method with high EC prediction performance and an efficient universal protein sequence embedding method, we proposed the DMLF approach, composed of an embedding core and a learning core. These two cores operate relatively independently. The embedding core is responsible for embedding protein sequences into a machine-readable matrix. The learning core is responsible for solving specific downstream biological tasks (e.g., enzyme and non-enzyme prediction, multifunctional enzyme prediction, and EC number prediction). The overall scheme of DMLF is illustrated in Fig. \ref{fig:framework}.

\noindent $\blacksquare$ \textbf{Core 1: Embedding }

The objective of this core is to calculate the embedding representations for protein sequences. For protein sequence encoding/embedding, recent studies have shown the superior performance of deep learning-based methods compared to traditional methods \cite{anteghini2021exploiting, martiny2021deep}. Accordingly, we only compared one-hot encoding to show the difference between these two kinds of embedding in this study. Here, we adopted three different embedding methods to calculate the sequence embedding patterns that adequately represent protein sequences. The first one is commonly the used one-hot encoding \cite{elabd2020amino}. The second is Unirep \cite{alley2019unified}, an mLSTM "babbler" deep representation learner for proteins. We used the last layer for protein representation. The third is the evolutionary scale modeling embedding method (ESM) \cite{rao2020transformer}, a pretrained transformer language model for protein representation. We used the hidden states from the 1st, 32nd, 33rd layers as protein embeddings.

\noindent $\blacksquare$ \textbf{Core 2: Learning }

The learning core is specialized to perform specific biological tasks using different agents. In this work, the learning core includes three agents. Agent-1 is a binary classifier that performs enzyme or non-enzyme prediction. This classifier was constructed using KNN \cite{zhang2017learning}. Agent-2 is a multi-classifier that predicts the number of putative functions for a given enzyme. It was implemented using an integrated sequence aligner, a gradient boost decision tree, and XGBoost. Agent-3 is also a multi-classifier that performs the EC number prediction task. As EC number prediction is an extreme multilabel classification (5852 classes in this benchmark), the performance of traditional multilabel classification methods such XGBoost, decision tree, and SVM is abysmal (less than 5\% in terms of accuracy). Therefore, we trained a scalable linear extreme classifier (SLICE)\cite{jain2019slice} to obtain a more reliable classification performance in this study. The details of agent implementation and parameter settings can be found in \href{./supplementary.pdf}{Supplemental, SI Appendix Materials and Methods C. Models}.

\noindent $\blacksquare$ \textbf{Integration, fine-tuning and output}
As illustrated in Fig. \ref{fig:framework}, the final EC number prediction output is an integrated process. As shown in Eq. \ref{eq:s1}, we formulated this integrated process as an optimization problem:
\vspace{-20pt}

\begin{equation}
  \centering
  \label{eq:s1}
  \mathop{MAX}\limits_{F1}\{f(ag_1, ag_2, ag_3, sa)\}
\end{equation}
where $ag_1$, $ag_2$, and $ag_3$ are the respective prediction results from Agent-1, Agent-2, and Agent-3, while $sa$ is the predicted result from multiple sequence alignment. The integration and fine-tuning process aims to maximize the optimizing objective. In this work, the objective is the performance of EC number prediction in terms of the F1 score. We used a greedy strategy to perform this optimization.

\subsection{Compared Baselines}

To evaluate our proposed method comprehensively, we compared our proposed method with four existing state-of-the-art techniques with 'GOOD' usability (see \href{./supplementary.pdf}{Supplemental, SI RELATED WORK}). Four state-of-the-art techniques are: CatFam, PRIAM (version 2), ECPred, and DeepEC.

\subsection{Evaluation Metrics}

To comprehensively evaluate the proposed method and existing baselines, we use 5 metrics to evaluate binary classification problems and 4 metrics to evaluate multiple classification problems. For the binary classification task, the evaluation criteria include ACC(accuracy),  PPV (positive predictive value,  precision), NPV(negative predictive value), RC (recall), and F1 value:

\begin{equation}
\label{lb:accuracy}
ACC = \frac {TP + TN } {TP+FP + TN+FN + UP + UN}
\end{equation}

\begin{equation}
\label{lb:precision}
PPV = \frac {TP} {TP+FP}
\end{equation}

\begin{equation}
\label{lb:npv}
NPV = \frac {TN} {TN+FN}
\end{equation}

\begin{equation}
\label{lb:recall}
Recall = \frac {TP} {TP+FN+UP}
\end{equation}

\begin{equation}
\label{lb:f1}
F1 = \frac {2 \times PPV \times Recall} {PPV+Recall}
\end{equation}
where $TP$ is the true positive value that represents the number of samples correctly identified as positive, $FP$ is the false positive value that represents the number of samples wrongly identified as positive, $TN$ is the true negative value that represents the number of samples correctly identified as negative, $FN$ is false negative value that represents the number of samples wrongly identified as negative, $UP$ is unclassified positive samples, and $UN$ is unclassified negative samples.

For multiple classification problems, the evaluation criteria included mACC (macro-average accuracy), mPR(macro-average precision), mRecall(macro-average recall), and mF1(macro-average F1 value): 

\begin{equation}
\label{lb:macc}
mACC= \frac{\sum_{i=1}^{n}{ACC_i}} {n}, {n = 1, 2, 3, \cdots, N }
\end{equation}

\begin{equation}
\label{lb:mpr}
mPR= \frac{\sum_{i=1}^{n}{PPV_i}} {n}, {n = 1, 2, 3, \cdots, N }
\end{equation}

\begin{equation}
\label{lb:mrecall}
mRecall= \frac{\sum_{i=1}^{n}{Recall_i}} {n}, {n = 1, 2, 3, \cdots, N }
\end{equation}

\begin{equation}
\label{lb:mf1}
mF1= \frac{2 \times mPR \times mRecall} {mPR \times mRecall}, {n = 1, 2, 3, \cdots, N }
\end{equation}
where $N$ represents the total number of classes, while $ACC_i$, $PPV _i$, and $Recall_i$ represent the accuracy, precision, and recall of the $i$-th class in a one-VS-all mode\cite{rifkin2004defense}, respectively.

\section{Results}
\subsection{Embedding Core Performance Evaluation}

\begin{figure}[htbp]
    \centering	
	\includegraphics[width=0.80\textwidth]{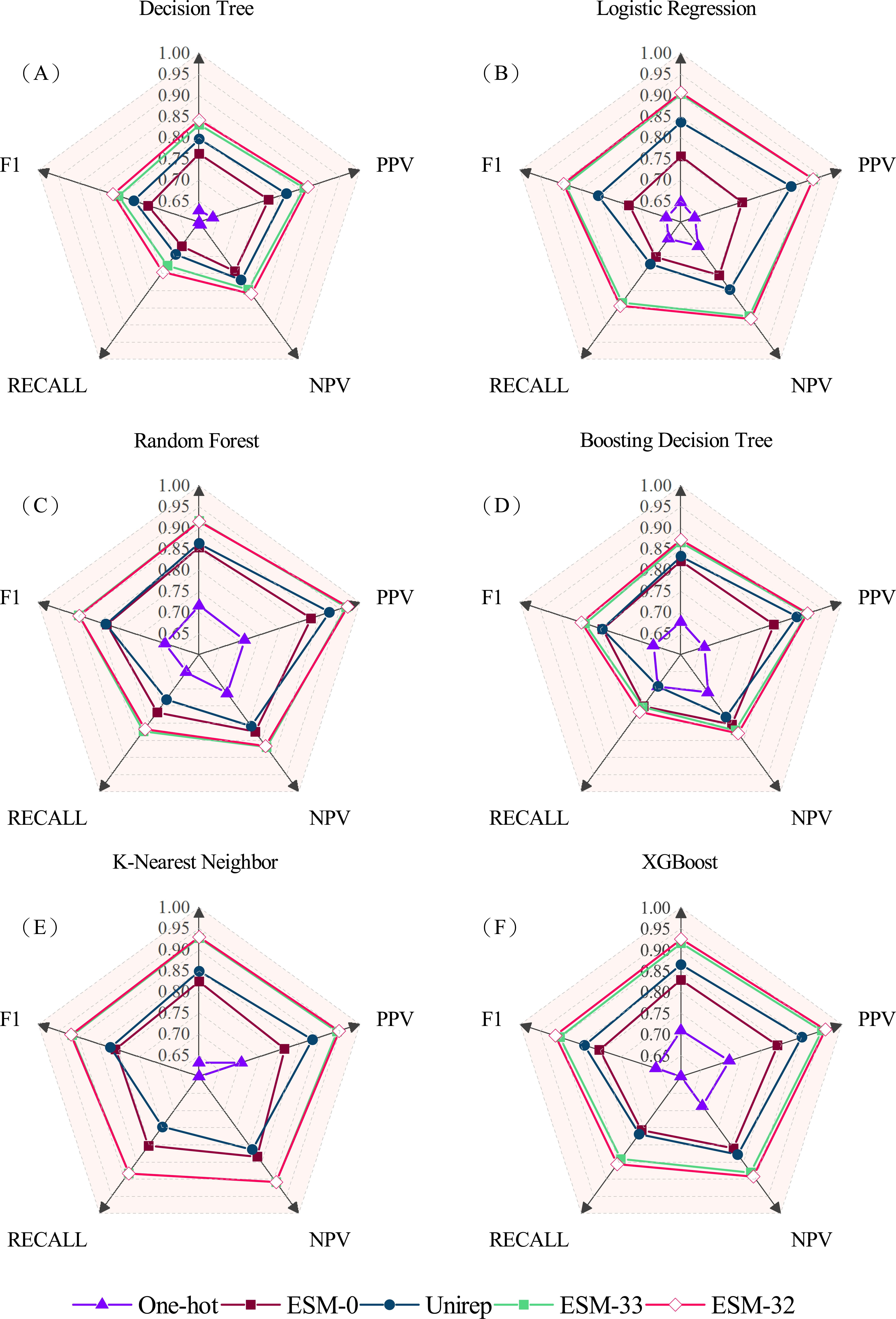}
	\caption{
        Performance comparison of different embedding methods for enzyme or non-enzyme annotation. 
	}
	\label{fig:embedding_performance}
\end{figure}

We evaluated five different protein embedding methods, one-hot embedding, Unirep embedding, and ESM embedding with three different layers (0,32,33) in our three proposed tasks. We used six machine learning baselines, including K-nearest neighbor (KNN), logistic regression (LR), XGBoost, decision tree (DT), random forest (RF), and gradient boosting decision tree (GBDT) to conduct this evaluation. For embedding, ESM-32 exhibited the best overall performance among all six baselines regarding all evaluation metrics for embedding (see Supplemental Tables S12 and S13). As shown in Fig. 3, in task 1, ESM-32 achieved 21.67 and 6.03\% improvements over one-hot and Unirep in terms of accuracy, as well as 27.20 and 7.32\% in terms of F1, respectively \href{./supplementary.pdf}{(see Supplemental Table S11)}. This experiment suggests that better embedding can lead to better learning performance, and deep latent representation can comprehensively represent the protein sequence. The embedding performance of ESM-32 was better than that of ESM-33, suggesting that a deeper embedding layer is not always better. DMLF can automatically choose the best embedding methods based on the downstream tasks, and ESM-32 exhibited the best performance in this work. 

\subsection{Task 1: Enzyme or Non-enzyme Prediction}

In this work, the workflow of enzyme number assignment is: \textit{task 1}, determine whether the given protein sequence is an enzyme $\overrightarrow{}$ {task 2}, if the given protein is an enzyme, then predict how many enzyme functions it can perform; $\overrightarrow{}$ \textit{task3}, assign an EC number for each enzyme function. According to this workflow, the first benchmarking task is enzyme or non-enzyme prediction. In this task, we trained an integrated binary classification model, which is driven by KNN and sequence alignment. KNN was implemented using scikit-learn, and the alignment was implemented using diamond v2.0.11.

\begin{figure}[ht]
    \centering	
	\includegraphics[width=0.8\textwidth]{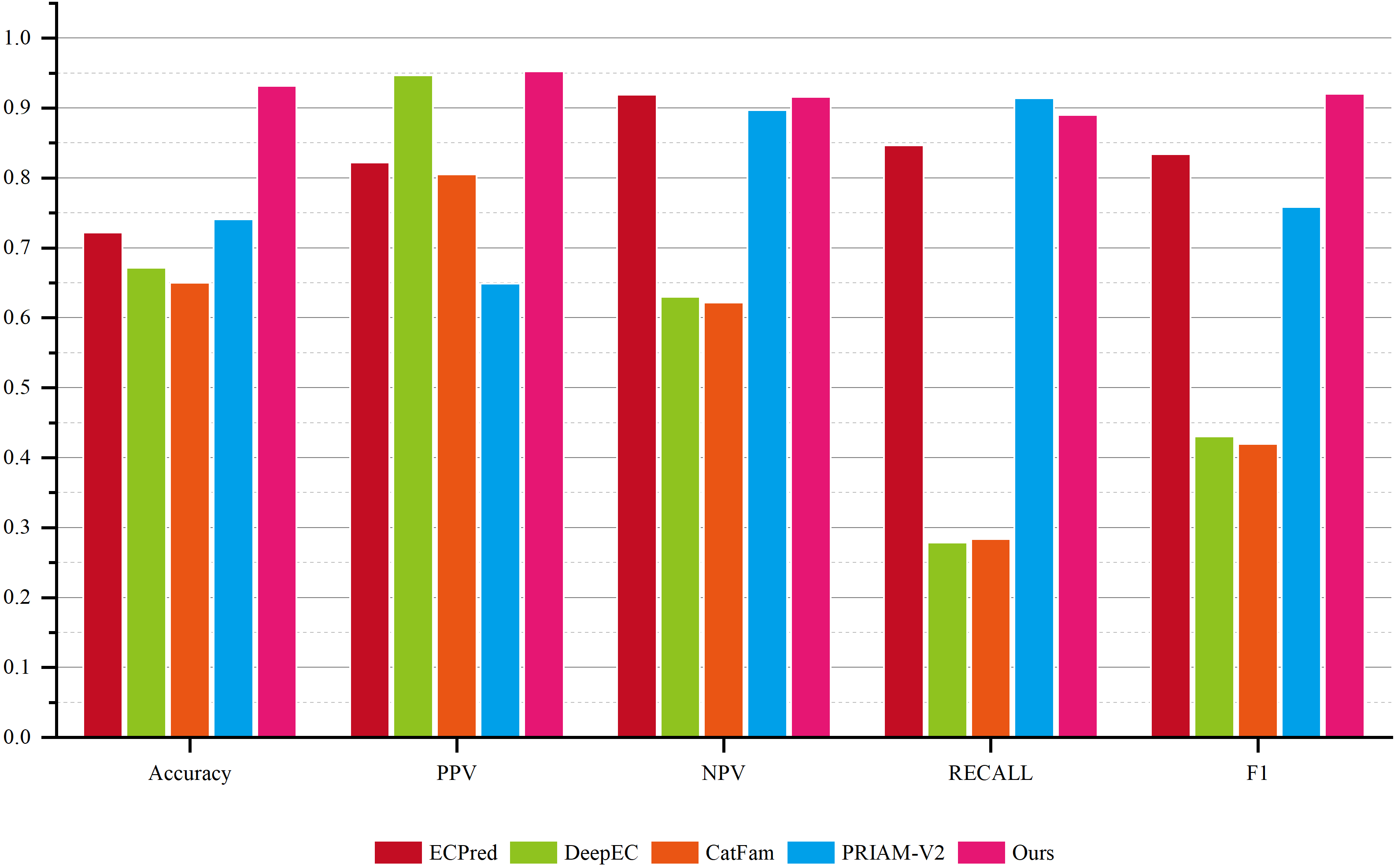}
	\caption{
        Task 1 Comparison of enzyme or non-enzyme annotation  
	}
	\label{fig:task1_performance}
\end{figure}

As shown in Fig. \ref{fig:task1_performance}, our method can achieve scores of 93.12, 95.25, and 88.99\% in terms of accuracy, precision, and recall, respectively \href{./supplementary.pdf}{(see Supplemental Table S8)}. Compared with other state-of-the-art tools and techniques the overall accuracy was greatly improved. For example, DeepEC yielded 74.10\%, compared with 93.12\% using our algorithm. Many previous methods were designed to obtain high precision while neglecting accuracy, NPV, and recall. For example, DeepEC can reach 94.68\% precision while recall is only 20.83\%. Methods that only offer high precision are very likely to miss many new functions. The F1 score might be a better evaluation metric for the EC assignment of real-world proteins.

\subsection{Task 2: Multifunctional Enzyme Prediction}
The second benchmarking task we addressed is multifunctional enzyme prediction. The backward prediction engine is agent 2 (see Fig. \ref{fig:framework}). In this task, we trained an integrated multiple-classification model driven by sequence alignment and XGBoost.

\begin{figure}[ht]
    \centering	
	\includegraphics[width=0.8\textwidth]{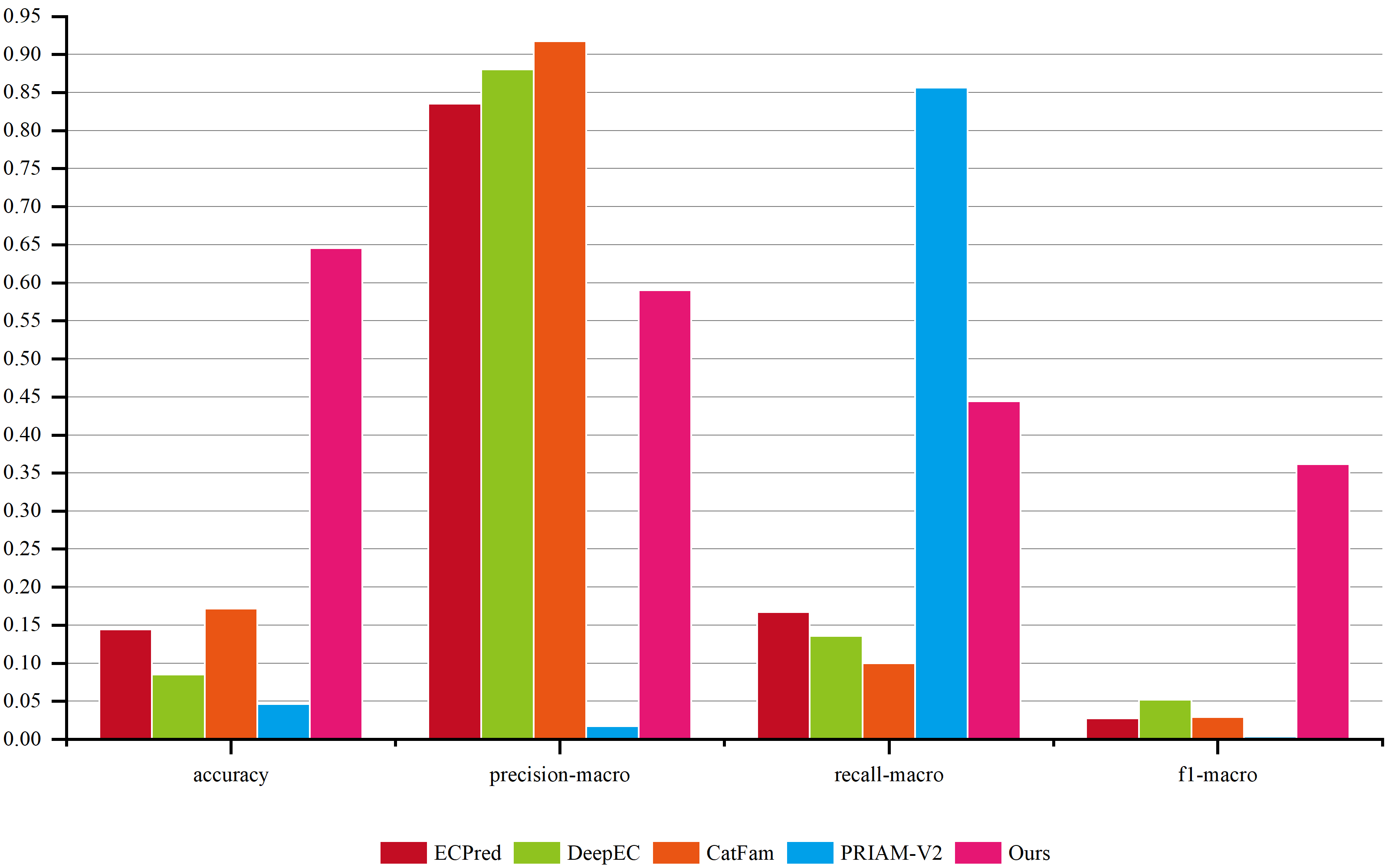}
	\caption{
        Task 2, Comparison of multifunctional enzyme prediction results. 
	}
	\label{fig:task2_performance}
\end{figure}

As shown in Fig. \ref{fig:task2_performance}, our method was superior to existing baselines \href{./supplementary.pdf}{(see Supplemental Table S9)}. For example, the accuracy and recall of DeepEC were 8.52 and 13.6\%, respectively. Moreover, the f1-macro of ECPred, DeepEC, CatFam, and PRIAM-V2 was less than 6\%, lower than a random prediction accuracy of 10\%. Hence, the performance was notably insufficient when dealing with multifunctional enzyme prediction. The low performance is mainly due to a lack of multifunctional enzyme data (see Table \ref{tab:dataset_quantity}). Although our proposed method is 6.3 times better than random prediction in terms of accuracy, the performance is still insufficient, so it should be further improved in future work.

\subsection{Task 3: Enzyme Commission Number Prediction}

This task corresponds to agent-3 in DMLF. In order to develop a balanced EC number prediction algorithm with high accuracy combined with reasonable precision and recall, we trained an extreme multi-label classification model.

\begin{figure}[ht]
    \centering	
	\includegraphics[width=0.8\textwidth]{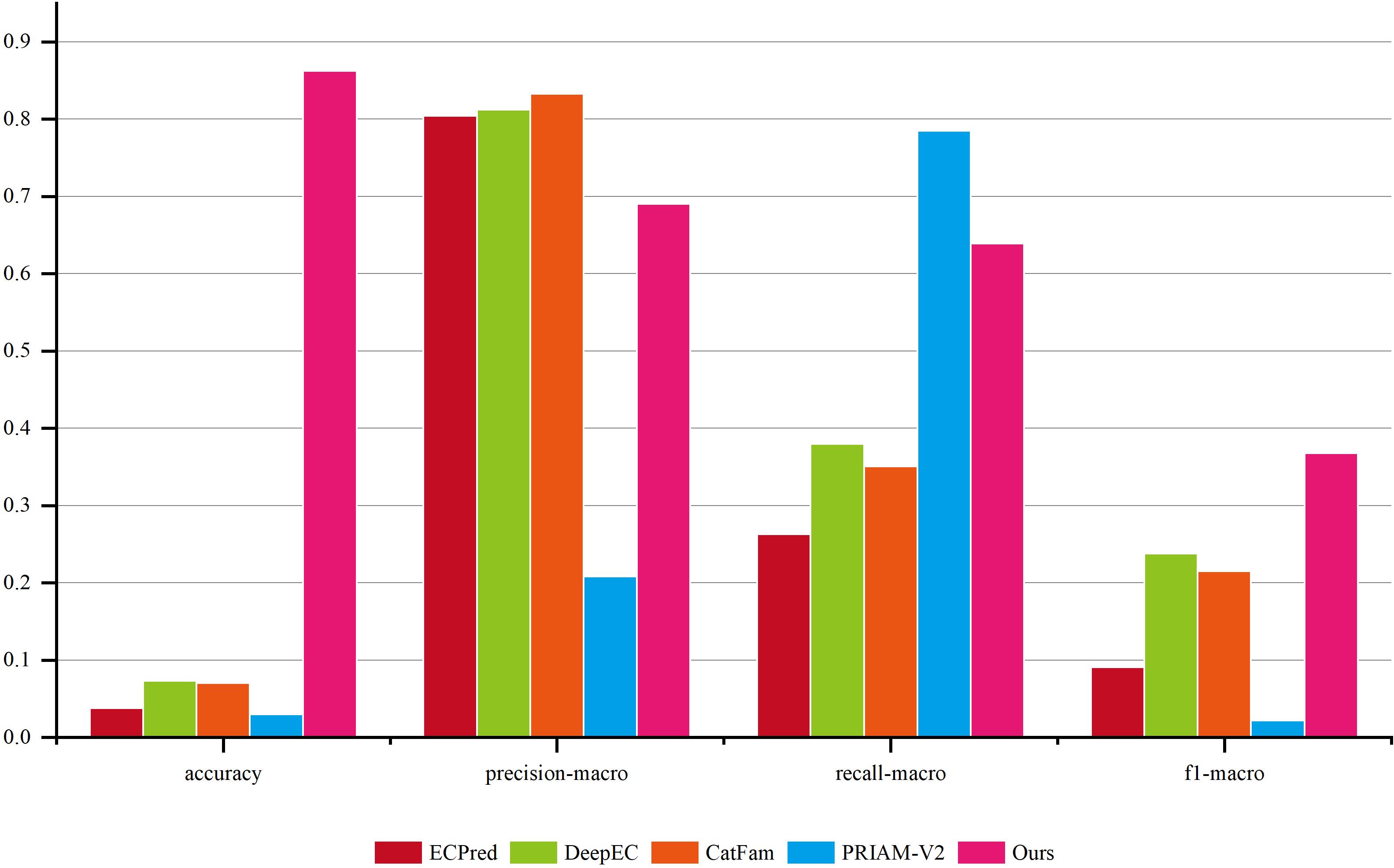}
	\caption{
        Task 3 Comparison of EC number prediction results. 
	}
	\label{fig:task3_performance}
\end{figure}

Our method achieved 86.91\% accuracy with 69\% precision and 63.88 recall (Fig. \ref{fig:task3_performance}), which means that if 100 protein sequences were uploaded for annotation, we can obtain approximately 87 correct annotations. PRIAM is mainly designed to include more sequences, so the recall is high (78.48\%), while the accuracy (3.0\%) and precision (20.80\%) are very low. DeepEC, ECPred, and CATFAM pursue high precision, so the accuracy is very low (less than 7.5\%), which means that if we upload 100 protein sequences for annotation, we can only obtain 7.5 correct annotations while the remaining 92.5 are wrong. Obviously, our method shows a clear advantage in terms of EC number assignment.

\section{Web Server Implementation}

\begin{figure}[ht]
    \centering	
	\includegraphics[width=0.8\textwidth]{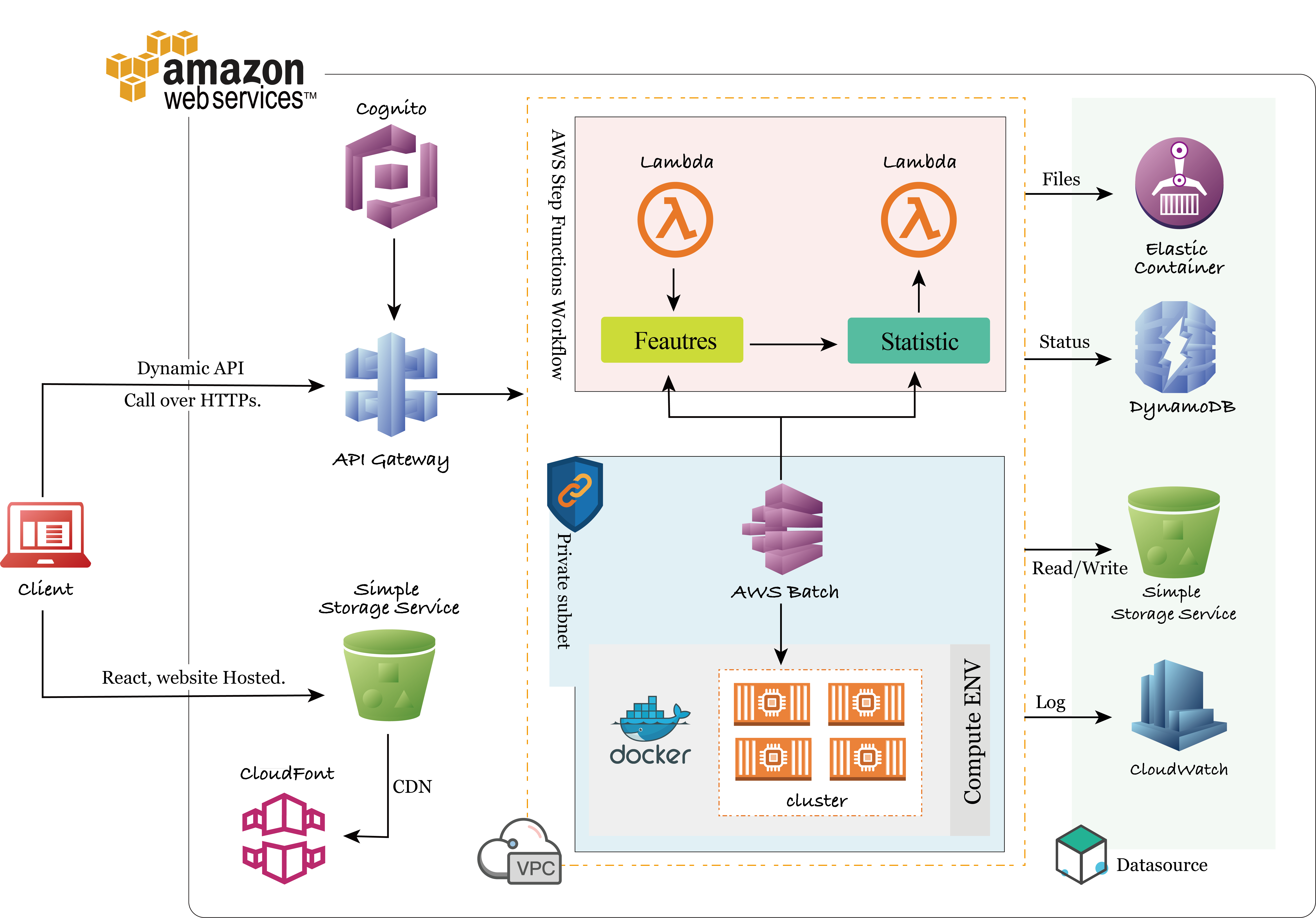}
	\caption{
        The architecture of the web platform. 
	}
	\label{fig:archi}
\end{figure}

To make the workflow accessible for biologists around the world, we built a web application (\href{https://ecrecer.biodesign.ac.cn/}{https://ecrecer.biodesign.ac.cn/}, Fig. \ref{fig:archi}). End-users can simply upload sequences to our platform, and then click the submit button to trigger the prediction workflow. In general, the whole workflow can be completed in a few seconds. We use Amazon DynamoDB to store job information, and users can track the previous submission records and corresponding status information. Once the analysis is finished, the user can view or download the corresponding results.

For EC assignment workflow, we use Amazon ECR to store Docker images, which packages a set of bioinformatics software, such as diamond and in-house python scripts. We built a scalable, elastic, and easily maintainable batch engine using AWS Batch. This solution took care of dynamically scaling our computer resources in response to the number of runnable jobs in our job queue. Finally, we used AWS step functions to coordinate the components of our applications easily, process message passed from AWS API Gateway, and invoke the workflows asynchronously. AWS API Gateway was used as the API server to handle the HTTP requests and route traffic to the correct backends. The static website was hosted by AWS S3 and sped up using AWS CloudFront.

\section{Case Study and Discussion}

When dealing with the EC assignment problem in a daily production scenario, ECRECer offers two optional modes for end-users: a prediction mode and a recommendation mode. In prediction mode, we provide the results with the highest probability, while in the recommendation mode, we deliver up to 20 possible EC number annotations ranked by their respective likelihood. Here, we present an up-to-date EC number prediction case to simulate the real-time challenge by conducting EC number assignments in the prediction mode. We collected testing protein sequences from June 2020 to November 2021, encompassing 1968 records. These data were not employed in the development process of the existing methods or our proposed method, which is in line with the daily production scenario. In this evaluation, we compared our method with the state-of-the-art method DeepEC. The comparison results demonstrated the exceptional performance of our proposed method in terms of accuracy and F1 score. Specifically, our method successfully predicted 1739 records, while DeepEC correctly predicted 1025 only records (88.36 vs. 52.08\%, 60.40 vs. 10.96\% in terms of accuracy and F1, respectively, 992 identical prediction tasks). For the sake of comparison with methods not handling the 7th EC class (translocases), we reconstructed a reduced sub-dataset comprising 997 enzyme sequences labeled with EC codes from classes 1 to 6. In this scenario, our methods still deliver a better performance. ECRECer and DeepEC predicted 807 and 100 correct enzyme records, respectively (96 identical prediction tasks). This proved the outstanding performance of ECRECer in enzyme annotation.

\begin{figure}[ht]
    \centering	
	\includegraphics[width=0.8\textwidth]{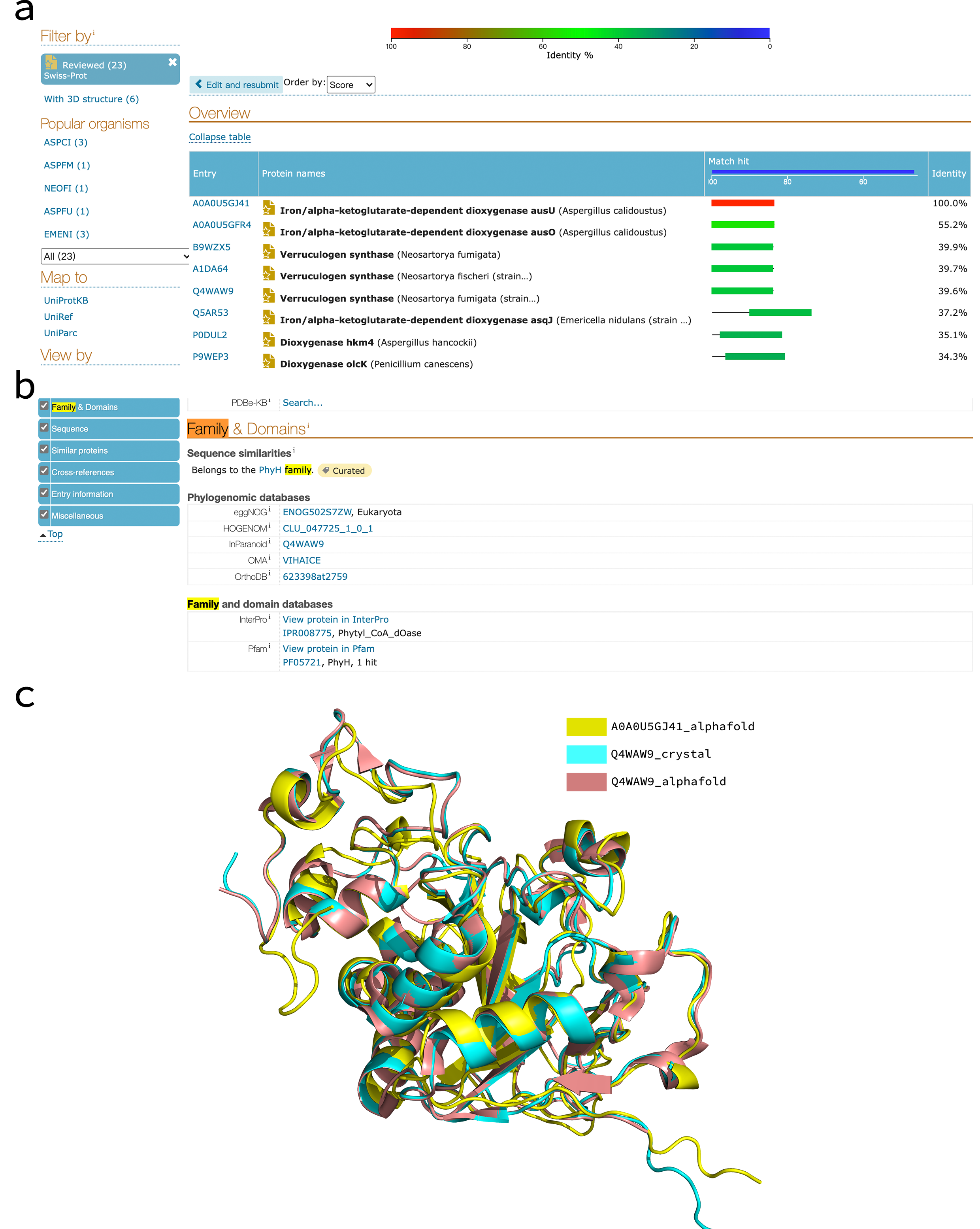}
	\caption{
        a) Blast search of the protein sequence against the UniProtKB database; b) Annotation of protein families and domains; c)Comparison of  structural similarity.
	}
	\label{fig:Figure8}
\end{figure}

We present another use case to show that our tools can be used for EC number completion. In the databases, many enzymes with EC numbers exist in an uncompleted three-level, two-level, or even one-level state. However, these proteins with incomplete EC numbers cannot directly be utilized for retrieving enzymatic reactions. Here, end-users can utilize ECRECer for EC number completion. For example, in the above case, we mis-predicted 163 monofunctional enzymes, 67 of which had incomplete EC numbers, while ECRECer completed 38 records to the final fourth level. For instance, for hybrid PKS-NRPS synthetase Phm1 (UniProt ID: A0A2Z5XAL7, EC: 2.3.1.-), ECRECer predicted the fourth-level EC number 2.3.1.41. Interestingly, after numerous literature reviews, we found that this protein has an active site based on PROSITE-ProRule annotation (PROSITE-ProRule: PRU10022), and it is a beta-ketoacyl synthases active site (\href{https://prosite.expasy.org/rule/PRU10022}{https://prosite.expasy.org/rule/PRU10022}). Therefore, it is very likely that this protein indeed has beta-ketoacyl-acyl-carrier-protein synthase activity. Another example is iron/alpha-ketoglutarate-dependent dioxygenase AusU (UniProt ID: A0A0U5GJ41). This protein has a two-level EC number in the database (1.14.-.-). When we used ECRECer for EC annotation, it assigned this protein with the fourth-level EC 1.14.11.38. This protein was recently integrated into UniProtKB/Swiss-Prot (September 29, 2021). After blasting it against the UniProtKB database, we found that the top 5 reviewed proteins with the highest identities include three verruculogen synthase (Fig. \ref{fig:Figure8}a). We took protein Q4WAW9 as an example, and found that both genes belong to exactly the same protein families with the same domains (Fig. \ref{fig:Figure8}b). To further validate the results, we compared the structure of A0A0U5GJ41 (alphfold2 predicted) and Q4WAW9 (alphfold2 predicted and crystal structure). The results showed that these two proteins have a highly similar structure (see Supplemental Figs. S3-S6) with small RMSD (1.104). Therefore, the protein could be potentially annotated as EC 1.14.11.38 as well.

In addition to EC number assignment, another advantage of ECRECer is the recommendation of EC numbers, which makes our tool unique. The recommendation is particularly helpful for the discovery of multifunctional enzymes. For example, in the case of annotating the intradiol ring-cleavage dioxygenase PrcA (UniProt ID: A2R1P9, EC: 1.13.11.3), the prediction mode indicates EC:1.13.11.1 as the most likely EC number, while the recommendation mode presents a list of recommendations [1.13.11.1, 1.13.11.3, 3.5.2.17, ... ]. The correct EC number is the second in this list. Actually, the substrates of the two reactions are very similar, and 3,4-dihydroxybenzoate only has an additional carboxyl group compared with catechol. Hence, PrcA might be a multifunctional enzyme. Another example is laccase (UniProt ID: A0A7T1FRB0, EC: 1.10.3.2), as its functional annotation is relatively sparce in the reviewed Swiss-Prot data. In this case, DeepEC, sequence alignment, and our prediction mode give out the prediction of a non-enzyme, but when we used the recommendation mode, we obtained a recommendation list [non-enzyme, 1.10.3.2, 1.-.-.-, 6.1.1.3, ...], and the correct annotation as offered as the second recommendation.  

To demonstrate the inclusiveness and predictive ability of our proposed method, we conducted EC number prediction on an unreviewed protein family. \textit{Corynebacterium glutamicum}, the famous industrial workhorse for amino acid production with a current output of over 6 million tons per year (Lee et al., 2016), is increasingly being adopted as a promising chassis for the biosynthesis of other compounds. However, unlike \textit{E. coli} (1652 protein sequences with EC numbers out of 4322 proteins, 38.2\%), the protein sequences of \textit{Corynebacterium glutamicum} were not well annotated. Out of 3305 protein sequences, only 537 were reviewed and included in the Swiss-Prot database (357 proteins have assigned EC numbers). We used the other 2768 protein sequences to compare our tool with DeepEC. Our approach was able to assign 1056 proteins with EC numbers s, while DeepEC only assigned 157 EC numbers (123 same EC numbers between DeepEC and ECRECer). Although there is no gold standard to decide which prediction is correct, we believe our algorithm should provide a more reasonable prediction as the proportion of protein sequences with EC numbers is similar to that of \textit{E. coli} (42\% vs. 15.5\% in the case of DeepEC). The newly predicted EC numbers for the protein sequences are crucial for further analysis, such as retrieving metabolic reactions for genome-scale modeling.

\section{Conclusion}
In this work, we proposed a novel dual-core multiagent learning framework to complete three benchmarking tasks: 1) enzyme or non-enzyme annotation; 2) predicting the number of catalytic functions of a single multifunctional enzyme; and 3) EC number prediction. The method developed in this work has two calculation cores, an embedding core and a learning core. The embedding core is responsible for selecting the best available embedding method among one-hot, Unirep, and ESM to calculate sequence embeddings. The learning core is responsible for completing the specific benchmarking tasks using the best calculated protein sequence embedding as input.

We were guided by two principles in the design of our methods. The first principle is high usability (both can be accessed via the world-wide-web and provide standalone suit for high throughput prediction) with relatively balanced prediction performance (which can achieve the best accuracy with reasonable precision and recall). The second principle is providing comprehensive evaluation metrics with accessible reproduction datasets and source codes. To implement the first principle, we proposed DMLF. To implement the second principle, we provided a web server, standalone packages and opened all the source codes, including data preprocessing, dataset buildup, model training, and model testing/evaluation.

Experiments on real-world datasets and comprehensive comparisons with existing state-of-the-art methods demonstrated that our tool is highly competitive, has the best performance with high usability, and meets the proposed objectives. Although our tool exhibited the best performance, it still has much space for improvement. For example, the performance of multifunctional enzyme annotation is relatively low, while the accuracy and recall of EC number annotation is less than 90\%. Our feature work will focus on improving the prediction precision.
\vspace{15pt}
\begin{mdframed}[leftmargin=-10pt,rightmargin=10pt]
    \vspace{5pt}
    \textbf{Key Points}
    \begin{itemize}
        \item A multiagent dual-core learning framework is proposed to predict Enzyme Commission (EC) Numbers by using protein sequence data.

        \item A protein language model and an extreme multi-label classifier are adopted to reduce the heavy head-crafted feature engineering and elevate the prediction performance.

        \item The proposed framework remarkably outperforms the existing state-of-the-the-art method in terms of accuracy and F1 score by 70\% and 20\%, respectively.

        \item An online service and an offline bundle are provided for end-users to annotate EC numbers in high-throughput easily and efficiently.
    \end{itemize}
    \vspace{2pt}
\end{mdframed}

\section{Supplementary Data}
Supplementary data are available onle at \url{https://github.com/kingstdio/ECRECer/blob/main/document/supplementary.pdf}

\section{Data availability}
The data underlying this article are available in the article and in its online Supplementary Material. The code of ECRECer,the training data,and the prediction results areavailable at url{https://ecrecer.biodesign.ac.cn/}.

\section{Author contributions statement}
Z.S. and X.L. designed and implemented the model, conducted the experiments, analyzed the results and wrote the manuscript. H.M. and Q.Y. reviewed the manuscript. R.W. and H.L. designed the website.

\section{Funding}
This work was supported by the National Key Research and Development Program of China [2018YFA0900300, 2020YFA0908300], the International Partnership Program of Chinese Academy of Sciences [153D31KYSB20170121], the Youth Innovation Promotion Association CAS, and Tianjin Synthetic Biotechnology Innovation Capacity Improvement Project [TSBICIP-PTJS-001, TSBICIP-CXRC-018].

\bibliographystyle{unsrt}

\begin{thebibliography}{10}

    \bibitem{uniprot2021uniprot}
    The~UniProt Consortium.
    \newblock Uniprot: the universal protein knowledgebase in 2021.
    \newblock {\em Nucleic Acids Research}, 49(D1):D480--D489, 2021.
    
    \bibitem{ryu2019deep}
    Jae~Yong Ryu, Hyun~Uk Kim, and Sang~Yup Lee.
    \newblock Deep learning enables high-quality and high-throughput prediction of
      enzyme commission numbers.
    \newblock {\em Proceedings of the National Academy of Sciences},
      116(28):13996--14001, 2019.
    
    \bibitem{furnham2009missing}
    Nicholas Furnham, John~S Garavelli, Rolf Apweiler, and Janet~M Thornton.
    \newblock Missing in action: enzyme functional annotations in biological
      databases.
    \newblock {\em Nature chemical biology}, 5(8):521--525, 2009.
    
    \bibitem{ecmuberwiki2021}
    Wikimedia.
    \newblock Enzyme commission number.
    \newblock [EB/OL], 2021.
    \newblock \url{https://en.wikipedia.org/wiki/Enzyme_Commission_number} Accessed
      November 29, 2021.
    
    \bibitem{hung2016sequence}
    Jui-Hung Hung and Zhiping Weng.
    \newblock Sequence alignment and homology search with blast and clustalw.
    \newblock {\em Cold Spring Harbor Protocols}, 2016(11):pdb--prot093088, 2016.
    
    \bibitem{yu2009genome}
    Chenggang Yu, Nela Zavaljevski, Valmik Desai, and Jaques Reifman.
    \newblock Genome-wide enzyme annotation with precision control: Catalytic
      families (catfam) databases.
    \newblock {\em Proteins: Structure, Function, and Bioinformatics},
      74(2):449--460, 2009.
    
    \bibitem{claudel2003enzyme}
    Clotilde Claudel-Renard, Claude Chevalet, Thomas Faraut, and Daniel Kahn.
    \newblock Enzyme-specific profiles for genome annotation: Priam.
    \newblock {\em Nucleic acids research}, 31(22):6633--6639, 2003.
    
    \bibitem{nursimulu2018improved}
    Nirvana Nursimulu, Leon~L Xu, James~D Wasmuth, Ivan Krukov, and John Parkinson.
    \newblock Improved enzyme annotation with ec-specific cutoffs using detect v2.
    \newblock {\em Bioinformatics}, 34(19):3393--3395, 2018.
    
    \bibitem{zhang2017cofactor}
    Chengxin Zhang, Peter~L Freddolino, and Yang Zhang.
    \newblock Cofactor: improved protein function prediction by combining
      structure, sequence and protein--protein interaction information.
    \newblock {\em Nucleic acids research}, 45(W1):W291--W299, 2017.
    
    \bibitem{li2016svm}
    Ying~Hong Li, Jing~Yu Xu, Lin Tao, Xiao~Feng Li, Shuang Li, Xian Zeng,
      Shang~Ying Chen, Peng Zhang, Chu Qin, Cheng Zhang, et~al.
    \newblock Svm-prot 2016: a web-server for machine learning prediction of
      protein functional families from sequence irrespective of similarity.
    \newblock {\em PloS one}, 11(8):e0155290, 2016.
    
    \bibitem{dalkiran2018ecpred}
    Alperen Dalkiran, Ahmet~Sureyya Rifaioglu, Maria~Jesus Martin, Rengul
      Cetin-Atalay, Volkan Atalay, and Tunca Do{\u{g}}an.
    \newblock Ecpred: a tool for the prediction of the enzymatic functions of
      protein sequences based on the ec nomenclature.
    \newblock {\em BMC bioinformatics}, 19(1):1--13, 2018.
    
    \bibitem{arakaki2009eficaz}
    Adrian~K Arakaki, Ying Huang, and Jeffrey Skolnick.
    \newblock Eficaz 2: enzyme function inference by a combined approach enhanced
      by machine learning.
    \newblock {\em BMC bioinformatics}, 10(1):1--15, 2009.
    
    \bibitem{akinosho2020deep}
    Taofeek~D Akinosho, Lukumon~O Oyedele, Muhammad Bilal, Anuoluwapo~O Ajayi,
      Manuel~Davila Delgado, Olugbenga~O Akinade, and Ashraf~A Ahmed.
    \newblock Deep learning in the construction industry: A review of present
      status and future innovations.
    \newblock {\em Journal of Building Engineering}, page 101827, 2020.
    
    \bibitem{li2020modern}
    Haoyang Li, Shuye Tian, Yu~Li, Qiming Fang, Renbo Tan, Yijie Pan, Chao Huang,
      Ying Xu, and Xin Gao.
    \newblock Modern deep learning in bioinformatics.
    \newblock {\em Journal of molecular cell biology}, 12(11):823--827, 2020.
    
    \bibitem{li2021deep}
    Yuxi Li, Yue Zuo, Houbing Song, and Zhihan Lv.
    \newblock Deep learning in security of internet of things.
    \newblock {\em IEEE Internet of Things Journal}, 2021.
    
    \bibitem{shi2021deep}
    Zhenkun Shi, Sen Wang, Lin Yue, Lixin Pang, Xianglin Zuo, Wanli Zuo, and Xue
      Li.
    \newblock Deep dynamic imputation of clinical time series for mortality
      prediction.
    \newblock {\em Information Sciences}, 579:607--622, 2021.
    
    \bibitem{shen2007ezypred}
    Hong-Bin Shen and Kuo-Chen Chou.
    \newblock Ezypred: a top--down approach for predicting enzyme functional
      classes and subclasses.
    \newblock {\em Biochemical and biophysical research communications},
      364(1):53--59, 2007.
    
    \bibitem{li2018deepre}
    Yu~Li, Sheng Wang, Ramzan Umarov, Bingqing Xie, Ming Fan, Lihua Li, and Xin
      Gao.
    \newblock Deepre: sequence-based enzyme ec number prediction by deep learning.
    \newblock {\em Bioinformatics}, 34(5):760--769, 2018.
    
    \bibitem{An2019}
    Ji~Yong An, Yong Zhou, Yu~Jun Zhao, and Zi~Ji Yan.
    \newblock {An Efficient Feature Extraction Technique Based on Local Coding PSSM
      and Multifeatures Fusion for Predicting Protein-Protein Interactions}.
    \newblock {\em Evolutionary Bioinformatics}, 15, 2019.
    
    \bibitem{Yang2018}
    Kevin~K. Yang, Zachary Wu, Claire~N. Bedbrook, and Frances~H. Arnold.
    \newblock {Learned protein embeddings for machine learning}.
    \newblock {\em Bioinformatics}, 34(15):2642--2648, 2018.
    
    \bibitem{alley2019unified}
    Ethan~C Alley, Grigory Khimulya, Surojit Biswas, Mohammed AlQuraishi, and
      George~M Church.
    \newblock Unified rational protein engineering with sequence-based deep
      representation learning.
    \newblock {\em Nature methods}, 16(12):1315--1322, 2019.
    
    \bibitem{rao2020transformer}
    Roshan Rao, Joshua Meier, Tom Sercu, Sergey Ovchinnikov, and Alexander Rives.
    \newblock Transformer protein language models are unsupervised structure
      learners.
    \newblock In {\em International Conference on Learning Representations}, 2020.
    
    \bibitem{anteghini2021exploiting}
    Marco Anteghini, Vitor Martins~dos Santos, and Edoardo Saccenti.
    \newblock In-pero: Exploiting deep learning embeddings of protein sequences to
      predict the localisation of peroxisomal proteins.
    \newblock {\em International Journal of Molecular Sciences}, 22(12):6409, 2021.
    
    \bibitem{martiny2021deep}
    Hannah-Marie Martiny, Jose Juan~Almagro Armenteros, Alexander~Rosenberg
      Johansen, Jesper Salomon, and Henrik Nielsen.
    \newblock Deep protein representations enable recombinant protein expression
      prediction.
    \newblock {\em bioRxiv}, 2021.
    
    \bibitem{elabd2020amino}
    Hesham ElAbd, Yana Bromberg, Adrienne Hoarfrost, Tobias Lenz, Andre Franke, and
      Mareike Wendorff.
    \newblock Amino acid encoding for deep learning applications.
    \newblock {\em BMC bioinformatics}, 21(1):1--14, 2020.
    
    \bibitem{zhang2017learning}
    Shichao Zhang, Xuelong Li, Ming Zong, Xiaofeng Zhu, and Debo Cheng.
    \newblock Learning k for knn classification.
    \newblock {\em ACM Transactions on Intelligent Systems and Technology (TIST)},
      8(3):1--19, 2017.
    
    \bibitem{jain2019slice}
    Himanshu Jain, Venkatesh Balasubramanian, Bhanu Chunduri, and Manik Varma.
    \newblock Slice: Scalable linear extreme classifiers trained on 100 million
      labels for related searches.
    \newblock In {\em Proceedings of the Twelfth ACM International Conference on
      Web Search and Data Mining}, pages 528--536, 2019.
    
    \bibitem{rifkin2004defense}
    Ryan Rifkin and Aldebaro Klautau.
    \newblock In defense of one-vs-all classification.
    \newblock {\em The Journal of Machine Learning Research}, 5:101--141, 2004.
    
    \end{thebibliography}

\end{document}


\vspace*{0.35in}
\title{\textit{\huge \textcolor[rgb]{0.11,0.11,0.70}{Supplemental Document}}\\ \text{ \huge Enzyme Commission Number Recommendation and} \\ \vspace{-8pt} \text{ \huge  Benchmarking based on Multi-agent Dual-core Learning }}

\begin{flushleft}
{   \large
    \textbf\newline{$ \bigstar  $ \underline{ Supplemental Document} }
    \vspace{25pt}
}
{\Large \textbf\newline{Enzyme Commission Number Recommendation and Benchmarking based on Multi-agent Dual-core Learning}}
\newline
\\
Zhenkun Shi\textsuperscript{1, 2, *},
Qianqian Yuan\textsuperscript{1, 2},
Ruoyu Wang\textsuperscript{1, 2},
Haoran Li\textsuperscript{1, 2},
Xiaoping Liao\textsuperscript{1, 2, (\Letter)},
Hongwu Ma\textsuperscript{1, 2, (\Letter)}
\\
\bigskip
$^1$Biodesign Center, Key Laboratory of Systems Microbial Technology, Tianjin Institute of Industrial Biotechnology, Chinese Academy of Sciences, 300308, Tianjin, China. \\
$^2$ National Technology Innovation Center of Synthetic Biology, 300308, Tianjin, China
\\
\bigskip
* zhenkun.shi@tib.cas.cn

\end{flushleft}

\section*{Abstract}
This is a supplementary document for the paper "Enzyme Commission Number Recommendation and Benchmarking based on Multi-agent Dual-core Learning". It provides details on preparing the data, selecting models, fine-tuning parameters, performance evaluation, as well as supplementary figures and tables that provide experimental details and support our conclusions. It also includes information on how to use the web service and offline bundles for high-throughput EC number prediction.

\linenumbers

\section{SI Related Work}
As EC number prediction is at the core of enzyme functional annotation, a large number of relevant computational techniques have been developed to assign EC numbers to unknown protein sequences. In this section, we will introduce seven of the most representative ones, ordered by their time of publication. The seven representative tools are listed in Table \ref{tab:baseline_usablity}. Next, we evaluated these tools based on the latest update time, distribution type (standalone packages, online web-service, or both), usability (’YES’ if it is available for use, ’NO’ if it is not available, ’Good’ if it can be used for high-throughput prediction) and citations of these tools up to 26 Aug 2021.

\begin{table}[htbp]
    \begin{adjustwidth}{-0.0in}{0in}
    \centering
    \caption{Usability of 7 EC prediction tools}
    \label{tab:baseline_usablity}
    \begin{tabular}{p{3cm}<{\centering}|p{2.0cm}<{\centering}|p{2.0cm}<{\centering}|p{2.0cm}<{\centering}|p{2.0cm}<{\centering}} \hline
    \rowcolor[HTML]{EFEFEF} 
    {Tools}        & {Last update}& {Type}     & {Usability} & {Citations}   \\  \hline
    CatFam$^1$     & 2009         & standalone & GOOD        & 71            \\ \hline
    SVMProt$^2$    & 2016         & online     & NO          & 88            \\ \hline
    PRIAM\_V2$^3$  & 2018         & both       & GOOD        & 365           \\ \hline
    DEEPre$^4$     & 2018         & online     & YES         & 132           \\ \hline
    ECPred$^5$     & 2018         & both       & GOOD        & 40            \\ \hline
    DEEPEC$^6$     & 2019         & standalone & GOOD        & 49            \\ \hline
    BENZ WS$^7$    & 2021         & online     & YES         & 0             \\ \hline
    \end{tabular}
      \begin{tablenotes}
        
        \item[1] 1. \href{http://www.bhsai.org/downloads/catfam.tar.gz}{http://www.bhsai.org/downloads/catfam.tar.gz}
        \item[2] 2. \href{http://bidd.group/cgi-bin/svmprot/svmprot.cgi}{http://bidd.group/cgi-bin/svmprot/svmprot.cgi}
        \item[3] 3. \href{http://priam.prabi.fr/REL_JAN18/index_jan18.html}{http://priam.prabi.fr/REL\_JAN18/index\_jan18.html}
        \item[4] 4. \href{http://www.cbrc.kaust.edu.sa/DEEPre/index.html}{http://www.cbrc.kaust.edu.sa/DEEPre/index.html}
        \item[5] 5. \href{https://ecpred.kansil.org/}{https://ecpred.kansil.org/}
        \item[6] 6. \href{https://bitbucket.org/kaistsystemsbiology/deepec/src/master/}{https://bitbucket.org/kaistsystemsbiology/deepec/src/master/}
            \item[7] 7. \href{https://benzdb.biocomp.unibo.it/}{https://benzdb.biocomp.unibo.it/}
      \end{tablenotes}
    \end{adjustwidth}
\end{table}

\subsection{CatFam}

CatFam \cite{yu2009genome} is a profile-controlled, sequence-based database that can be used to infer catalytic functions of proteins. CatFam uses an adjustable false positive rate to generate databases on-demand for different needs, such as functional annotation with different precision and hypothesis generation with moderate precision but better recall. CatFam uses profile-specific thresholds to ensure equal precision for each profile and ensure the best performance for all tasks. Comparison experiments were conducted based on three test sets and 13 bacterial genomes. The results demonstrated that CatFam outperforms PRIAM in terms of precision and coverage. CatFam has been developed for more than 12 years. Although the precision is not as good as in the latest ones, the recall remains good, and its code is still available. We therefore used CatFam as one of our baselines in this work.

\subsection{SVM-Prot}

SVM-Prot V2016 \cite{li2016svm} is a machine-learning method that was first published in 2003 and then updated in 2016. SVM-Prot is supplementary for predicting diverse classes of proteins compared with distantly-related or homologous-related methods. SVM-Prot employs 13 manually curated physicochemical features of proteins as inputs, nine of which are from Pse-in-One \cite{liu2015pse}, while the remaining four are self-calculated, such as molecular weight and solubility. The algorithm then uses these features to train an integrated SVM, KNN, PNN, and Blast model, to predict the EC numbers for new proteins. Sensitivity, precision, and specificity are evaluated on an independent evaluation dataset, which demonstrated the outstanding performance of SVM-Prot. However, to train an SVM classifier the time complexity is $O(n^2p+n^3)$ \cite{abdiansah2015time}, which is extremely time-consuming. More importantly, the web service provided by SVM-Prot is no longer available, and they did not provide their code for reimplementation and evaluation. Hence, the usability of SVM-Prot is weak.

\subsection{PRIAM-V2}

PRIAM V2 \cite{claudel2003enzyme} s a rules-based method for automated enzyme annotation with EC numbers proposed in 2003, with an updated version V2 published in 2018. It takes protein or nucleotide sequences as inputs and annotates them with EC numbers on an individual sequence level or a genome level. PRIAM utilizes a set of signatures composed of position-specific scoring matrices and patterns for sequence embedding, which is tailored for each enzyme entry to build its model. PRIAM uses the whole Swiss-Prot database to learn parameters and evaluate the method as well. The advantage of PRIAM is its high recall, and the code is available. Accordingly, we used PRIAM V2 as one of our baselines.

\subsection{DEEPre}

DEEPre \cite{li2018deepre} is a supervised end-to-end feature selection and classification model that uses a convolutional neural network (CNN) with a level-by-level strategy to predict enzyme functions. Unlike the above-mentioned method that needs manually curated features, DEEPre takes the raw sequence encoding as inputs, then extracts convolutional and sequential features from the raw encoding based on the classification result to directly boost the model performance. DEEPre is good at determining the main classes of enzymes on a separate low-homology dataset, while the performance is suboptimal when determining the fourth level EC numbers. DEEPre provides a webserver for the public but does not provide the source code for reimplementation and evaluation, and the webserver is not capable of high-throughput prediction. Thus, this algorithm is usable but not user-friendly.

\subsection{ECPred}

ECPred \cite{dalkiran2018ecpred} is a supervised hierarchical enzyme function prediction tool based on an ensemble of machine learning that can predict EC numbers to the fourth level. ECPred trains an independent model for each EC number level and uses three predictors, called SPMap, BLAST-kNN, and Pepstats-SVM, to integrate the output. ECPred was trained and validated using the enzyme entries located in the Swiss-Prot database. ECPred ingeniously constructed a positive set and a negative set to finely control the prediction performance. The experimental results showed its outstanding performance at level 0 EC number prediction. ECPred was published in late 2018. The most significant point of ECPred is its user-friendly workflow that provides both a web service, standalone packages, and the source code. Accordingly, we used ECPred as one of our baselines in this work.

\subsection{DEEPEC}

DeepEC \cite{ryu2019deep} is a deep learning method that enables high-quality and high-throughput prediction of EC numbers. DeepEC uses three CNN as its major engine and homology analysis as its supplementary engine to conduct EC number prediction. DeepEC predicts if the given amino sequence is an enzyme in the first CNN layer, and then specifies the third level of EC numbers in the second CNN layer, after which it assigns the fourth level in the final CNN layer. The primary objective of DeepEC is high precision, low computing time, and low disk space requirements. DeepEC is sensitive in detecting the effects of mutated domains/binding site residues. DeepEC did not provide a source code for self-training and reimplementation. It only provides well-trained parameters for local installation and prediction. However, no webserver is given. Considering its good performance in precision, we also used use DeepEC as one of our baselines in this work.

\subsection{BENZ WS}

BENZ WS \cite{baldazzi2021benz} is the latest published web service for four-level EC number annotation. It was first published in May 2021. BENZ WS filters a target sequence with a combined system of HMMs and PFAMs, after which it returns an associated four-level EC number if successful. BENZ WS can annotate both mono- and multifunctional enzymes. Compared with DEEPre and ECPred, BENZ WS is superior in terms of the true positive rate. However, the performance of BENZ WS is relatively inferior in terms of the false-negative rate. BENZ WS only provides a web interface to the end-user, so usability is given, but no source code or standalone suite is available, and the computational time is long. We therefore did not use BENZ WS as a baseline in this work.

\section{SI Appendix Materials and Methods}

\subsection{Preprocessing}
There are six steps (s0-s5) in data preprocessing: 
\begin{enumerate}
  \item [1)] remove the records with identical IDs, but changed sequences (updated sequences);
  \item [2)] for duplicated records, only keep one;
  \item [3)] make the EC numbers uniform and remove unnecessary spaces;
  \item [4)] based on the EC number, assign a unique label for each sequence;
  \item [5)] organize a uniform dictionary for EC label mapping;
  \item [6)] add enzyme catalytic function quantity labels to protein sequences.
\end{enumerate}

\subsection{Dataset}

A commonly used EC number prediction dataset is the EzyPred dataset from Shen and Zhou, published in 2007 \cite{shen2007ezypred}. The EzyPred dataset is a two-level EC number dataset that was extracted from the ENZYME database (released May 1, 2007), with a 40\% sequence similarity cutoff. This dataset contains 9,832 two-level specified enzymes and 9850 non-enzymes. The details of this dataset can be found in their published paper \cite{shen2007ezypred}. This dataset can only be used to predict two-level EC numbers, and the volume of this dataset is unsuitable for machine learning. Accordingly, the majority of the later studies used a similar approach to extract and construct datasets from Swiss-Prot \cite{ryu2019deep, li2018deepre}. The typical steps of constructing the dataset are as follows:

\begin{itemize}
  \item [{1)}] Obtain the latest reviewed protein data from Swiss-Prot and label the sequences as enzyme or none-enzyme utilizing the protein annotation.
  \item [{2)}] Exclude the multifunctional enzymes and those enzymes with incomplete EC number annotations.
  \item [{3)}] Exclude enzymes by sequence length, a typical threshold is $length \in [50, 50000]$.
  \item [{4)}] Use homology analysis tools to remove redundant sequences. The similarity threshold is manually determined, and a typical threshold is $40\%$.
  \item [{5)}] Randomly rearrange filtered enzyme data and randomly pick non-enzyme data with a similar size, then mix these data together as a standard dataset.
  \item [{6)}] Split the standard dataset into a training set and a testing set using a typical 8:2 ratio or split the standard dataset into a training set, validation set, and testing set using a typical 7:1:2 ratio.
\end{itemize}

However, these principles of dataset construction were explicitly designed for the EC number prediction of monofunctional enzymes and are not suitable for multifunctional enzymes. Moreover, the construction of training and testing datasets using randomly mixed data is not in accordance with the facts and may lead to information leaks. Beyond that, filtering sequences by length and homology may obscure patterns and other information, which will reduce the learning performance. Therefore, the steps of constructing the dataset in this work were more straightforward:

\begin{itemize}
  \item [{1*)}] Obtain the latest reviewed protein data from Swiss-Prot and label the sequences with three label vectors: enzyme or none-enzyme, monofunctional (labeled 1 or 0) or multifunctional enzyme (labeled with function counts), EC number (monofunctional enzymes have a single EC number, multifunctional enzymes have more than one EC number).
  \item [{2*)}] Rearrange the protein sequence order by annotation updated date. 
  \item [{3*)}] Use the latest three-year data as the testing set, while the rest is the training and validation set.
\end{itemize}

The implementation can be seen in chapter 5 and by referring to our source codes.

\subsubsection{Task 1 Enzyme and Non-enzyme Dataset}
Based on the above mentioned three principles, the enzyme and non-enzyme dataset (Table \ref{tab:dataset1}) uses the latest 3 years of Swiss-Prot data as the testing set, and the data before as the training set.
\begin{table}[htbp]
  \centering
  \caption{Description of the Enzyme and Non-enzyme Dataset}
  \label{tab:dataset1}
  \begin{tabular}{
                  p{3.7cm}<{\centering}|
                  p{3.7cm}<{\centering}|
                  p{3.7cm}<{\centering}
    }
  \hline \rowcolor[HTML]{EFEFEF} 
  ITEM       & Training set   & Testing set  \\ \hline
  Enzyme     & 222,567        & 3,304 \\ \hline
  Non-enzyme & 246,567        & 3,797 \\ \hline
  Total      & 469,134        & 7,101 \\ \hline
  \end{tabular}
\end{table}
\subsubsection{Task 2 Multifunctional Enzyme Dataset}
For the multifunctional enzyme prediction dataset, to minimize distractions from non-enzymes and balance the dataset, we excluded the non-enzyme data (Table \ref{tab:dataset2}). The remaining enzyme data was labeled based on the number of functions (i.e., 1, 2, ....,8). The details are listed below:

\begin{table}[ht]
  \begin{adjustwidth}{-0.6in}{0in}
  \centering
  \caption{Description of Multifunctional Enzyme Dataset}
  \label{tab:dataset2}
  \begin{tabular}{
                  p{2cm}<{\centering}|p{2cm}<{\centering}|
                  p{2cm}<{\centering}|p{2cm}<{\centering}|
                  p{2cm}<{\centering}|p{2cm}<{\centering}
                }
      \hline \rowcolor[HTML]{EFEFEF} 
        & \multicolumn{2}{c|}{Records} & 
        & \multicolumn{2}{c}{Records} \\ \cline{2-3} \cline{5-6}
      \rowcolor[HTML]{EFEFEF}   \multirow{-2}{*}{Functions}  & Trainning set & Testting set&  \multirow{-2}{*}{Functions}   & Training set & Testing set    \\ \hline
      1       & 210788   & 3052   & 5   & 206      & 6          \\ \hline
      2       & 9943     & 183    & 6   & 80       & 2          \\ \hline
      3       & 993      & 53     & 7   & 27       & 1          \\ \hline
      4       & 525      & 7      & 8   & 5        & 0          \\ \hline
  \end{tabular}
  \end{adjustwidth}
\end{table}

\subsubsection{Task 3 Enzyme Commission Number Dataset}
Following the three-datasets construction principle, the enzyme commission (EC) number dataset filtered the non-enzyme data after preprocessing. For a comprehensive and fair comparison with the state-of-the-art method DeepEC, we set the end-time of the training dataset to February 2018. This is because DeepEC only collected data before February 2018 for model training. If we use more recent data it will lead to an information leak problem. The dataset (Table \ref{tab:dataset3}) details are listed below:

\begin{table}[htbp]
  \begin{adjustwidth}{-0.6in}{0in}
  \centering
  \caption{Description of the Enzyme Commission Number Dataset}
  \label{tab:dataset3}
  \begin{tabular}{
                  p{5cm}<{\centering}|p{4cm}<{\centering}|
                  p{4cm}<{\centering}
                }
  \hline \rowcolor[HTML]{EFEFEF}
  Item                    & Train   & Test  \\ \hline
  Monofunctional          & 210,788 & 3,052 \\ \hline
  Multifunctional         & 11,779  & 252   \\ \hline
  Distinct EC   numbers   & 4,854   & 937   \\ \hline
  Incomplete EC   numbers & 209     & 128   \\ \hline
  Complete EC   numbers   & 4,645   & 809   \\ \hline
  Oxidoreductases         & 1,368   & 243   \\ \hline
  Transferases            & 1,433   & 258   \\ \hline
  Hydrolases              & 1,078   & 192   \\ \hline
  Lyases                  & 548     & 138   \\ \hline
  Isomerases              & 247     & 57    \\ \hline
  Ligases                 & 180     & 35    \\ \hline
  Translocases            & -       & 14    \\ \hline
  Set size                & 469,134 & 7,101 \\ \hline
  \end{tabular}
\end{adjustwidth}
\end{table}

\newpage

\section{Models}
\subsection{Agent 1. KNN model for enzyme or non-enzyme classification}

\begin{table}[htbp]
  \begin{adjustwidth}{-0.5in}{0in}
  \centering
  \caption{Performance comparison in enzyme or non-enzyme prediction}
  \label{tab:agent1_res}
  \begin{tabular}{c|c|c|c|c|c|c|c|c|c}
  \hline
  \rowcolor[HTML]{EFEFEF}
   &  &  &  &  &  & \multicolumn{4}{c}{Confusion Matrix} \\ \cline{7-10} 
  \rowcolor[HTML]{EFEFEF}
  \multirow{-2}{*}{Baseline}   &
  \multirow{-2}{*}{Accuracy}   &
  \multirow{-2}{*}{PPV}        &
  \multirow{-2}{*}{NPV}        & 
  \multirow{-2}{*}{Recall}     & 
  \multirow{-2}{*}{F1}         &TP      &FP       &FN     &TN   \\ \hline
  KNN     & \textbf{0.9248}   &0.9391   &\textbf{0.9134}  &\textbf{0.8965}   &\textbf{0.9173} &2962 & 192 & 342 & 3605 \\ \hline
  LR      & 0.9086            &0.9275   &0.8939  &0.8717   &0.8987 &2880 & 225 & 424 & 3572 \\ \hline
  XGBoost & 0.9214            &0.9499   &0.9000  &0.8774   &0.9122 &2899 & 153 & 405 & 3644 \\ \hline
  DT      & 0.8306            &0.8560   &0.8125  &0.7645   &0.8077 &2526 & 425 & 778 & 3372 \\ \hline
  RF      & 0.9086            &0.9614   &0.8726  &0.8372   &0.8950 &2766 & 111 & 538 & 3686 \\ \hline
  GBDT    & 0.8749            &0.9056   &0.8528  &0.8163   &0.8586 &2697 & 281 & 607 & 3516 \\ \hline
  \end{tabular}
\end{adjustwidth}
\end{table}

After a comprehensive evaluation using machine learning baselines (See Table \ref{tab:agent1_res}, below) , we adopted KNN as our first agent. The K Nearest Neighbor (KNN) method has been widely used in data mining and machine learning applications due to its simple implementation and distinguished performance \cite{zhang2017learning}. The optimized parameters are listed as follows:

\begin{lstlisting}
  # KNN optimized parameters
  KNeighborsParameters = {  n_neighbors = 5,
                            weights='distance',
                            algorithm = 'kd_tree',
                            leaf_size = '30',
                            p = 2,
                            metric = 'euclidean',
                            metric_params = None,
                            n_jobs = -2
                          }
\end{lstlisting}

\subsection{Agent 2. XGBoost model for multifunctional enzyme prediction}
As shown in Table \ref{tab:agent2_res}, when dealing with multifunctional enzyme prediction, the learning performance of KNN is not optimal. Thus, after a comprehensive evaluation, in agent 2, we choose XGBoost as our algorithm for multifunctional enzyme prediction. XGBoost is an implementation of gradient boosted decision trees that have shown superior performance in  many data science problems \cite{chen2016xgboost}. The optimized parameters used in this work are listed as follows:
\vspace{10pt}
\begin{lstlisting}
  # XGBoost optimized parameters
  XGBoostParameters = { objective = 'multi:softmax',
                        eval_metric='merror',
                        min_child_weight=6, 
                        max_depth=6, 
                        use_label_encoder=False,
                        n_estimators=120,
                        n_jobs = -2,
                        subsample = 0.5,
                        lambda = 1,
                        seed = 0
                      }
\end{lstlisting}

\begin{table}[htbp]
  \begin{adjustwidth}{-0.8in}{0in}
  \centering
  \caption{Performance comparison in multifunctional enzyme prediction}
  \label{tab:agent2_res}
  \begin{tabular}{p{2.5cm}<{\centering}|p{2cm}<{\centering}|p{3cm}<{\centering}|p{3cm}<{\centering}|p{2cm}<{\centering}}
  \hline
  \rowcolor[HTML]{EFEFEF}
  Basline     & Accuracy & Precision-Macro & Recall-Macro & F1-Macro \\ \hline
  KNN         & 0.8333   & 0.6786          & 0.6236       & 0.6239   \\ \hline
  LR          & 0.7619   & 0.7080          & 0.6379       & 0.5210   \\ \hline
  XGBoost     & 0.8492   & 0.8542          & 0.6307       & 0.6465   \\ \hline
  DT          & 0.7024   & 0.4791          & 0.4823       & 0.4800   \\ \hline
  RF          & 0.8532   & 0.8642          & 0.5465       & 0.5941   \\ \hline
  GBDT        & 0.8532   & 0.8566          & 0.5734       & 0.6023   \\ \hline
  \end{tabular}
\end{adjustwidth}
\end{table}

\subsection{Agent 3. SLICE model for EC number prediction}
As the number of classes reached 5852, the traditional machine learning methods could not achieve reasonable classification accuracy. Here, we introduced a scalable linear extreme classifier (SLICE) \cite{jain2019slice}. Based on the generative model and negative sampling techniques, SLICE uses approximate nearest neighbor search to learn low-dimensional dense features accurately and efficiently. As Slice is a one-vs-all algorithm, when dealing with the main class, the remaining (negative) samples from other classes result in an imbalance problem that greatly downgrades the classification performance. Considering the imbalance problem and the hierarchical information at different EC levels, we adopted a hierarchical navigable small-world graph \cite{malkov2018efficient} for downsampling the negative training samples in this work. The optimized parameters used in this work are listed as follows:
\vspace{50pt}

\begin{lstlisting}
  # Slice optimized parameters
  SliceParameters = { m = 100, 
                      efConstruction = 300, 
                      efSearch = 300,
                      k = 700,
                      C = 1,
                      f = 0.000001
                      iter = 1200,
                      type = 'L2R_L2LOss_SVC'
                    }
\end{lstlisting}

where $m$ is the maximum number of connections for each element per layer, $efConstruction$ is the size of the dynamic candidate in graph construction, $efSearch$ is the parameter to control the recall of the greedy search, $k$ is the cost co-efficient for linear classifiers, $f$ is the threshold value for sparsifying linear classifiers' trained weights to reduce model size. $iter$ is maximum iterations of algorithm for training linear classifiers.

\subsection{Integration, fine-tuning, and production}

\begin{figure}[ht]
  \centering
  \includegraphics[width=1.0\linewidth]{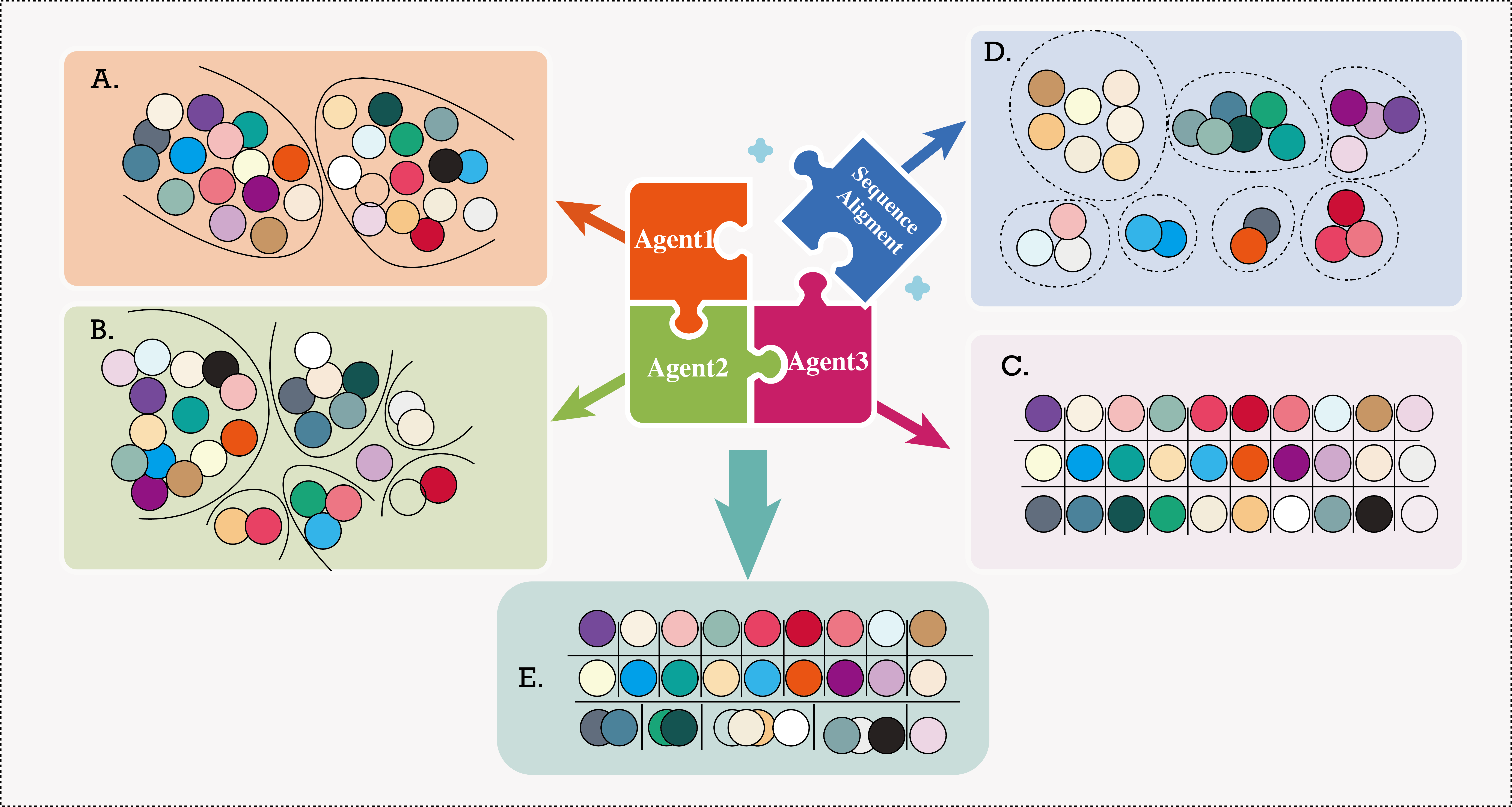}
  \caption{The integration and fine-tuning process before output.}
  \label{sifig:integration}
\end{figure}

As illustrated in Fig. \ref{sifig:integration}, the final EC number prediction output is an integrated process. As shown in SE. 1, we formulated this integrated process as an optimization problem:

\begin{align}
  \centering
  \label{eq:s1}
  \mathop{MAX}\limits_{F1}\{f(ag_1, ag_2, ag_3, sa)\}
  \tag{SE.1}
\end{align}
where $ag_1$, $ag_2$, and $ag_3$ are the predicted results from Agent1, Agent2, and Agent3, respectively, while sa is the predicted results from multiple sequence alignment. The integration and fine-tuning process aims to maximize the optimizing objective. In this work, the objective was the performance of EC number prediction in terms of the F1 score. We used a greedy strategy to finish this optimization.
    
\section{SI Appendix Figures}

  \begin{figure}[ht]
    \centering
    \includegraphics[width=1.0\linewidth]{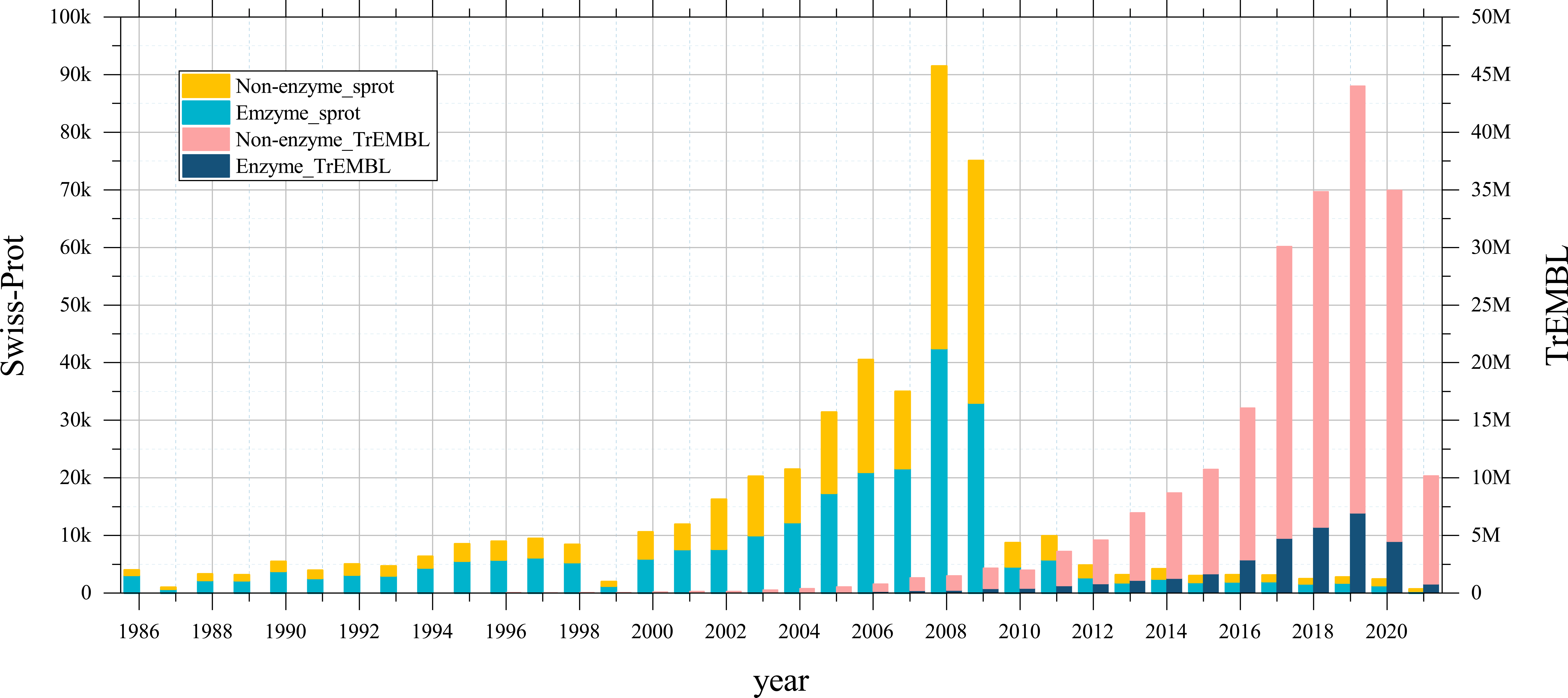}
    \caption{The number of records integrated into TrEMBL vs. Swiss-Prot since 1986.}
    \label{fig:false-color}
  \end{figure}

  \vspace{15pt}

  \begin{figure}[htbp!]
    \centering
    \includegraphics[width=0.85\linewidth]{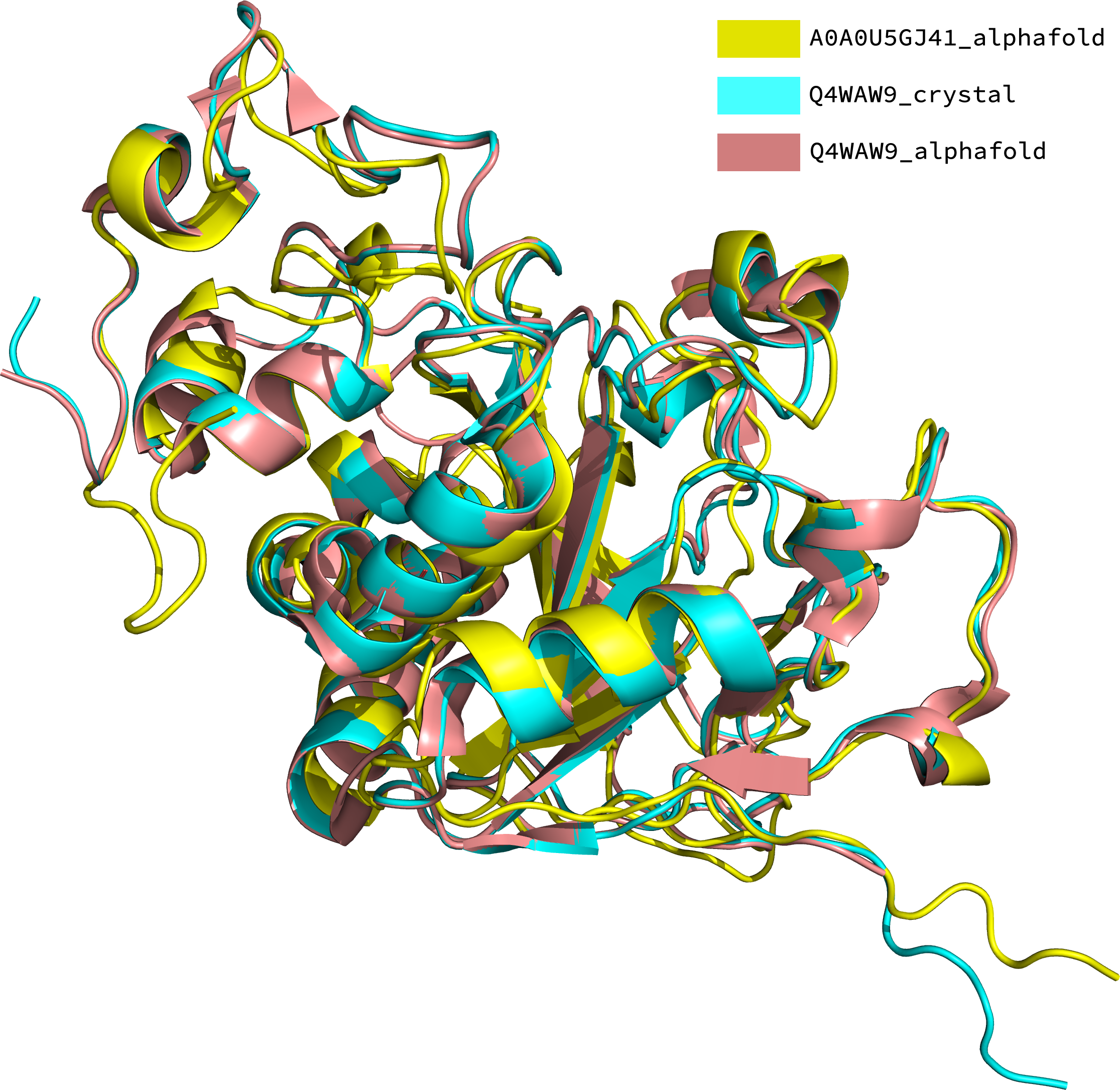}
    \caption{Structure alignment of proteins A0A0U5GJ41 and Q4WAW9.}
    \label{fig:aligned}
  \end{figure}

  \begin{figure}[htbp!]
    \centering
    \includegraphics[width=0.75\linewidth]{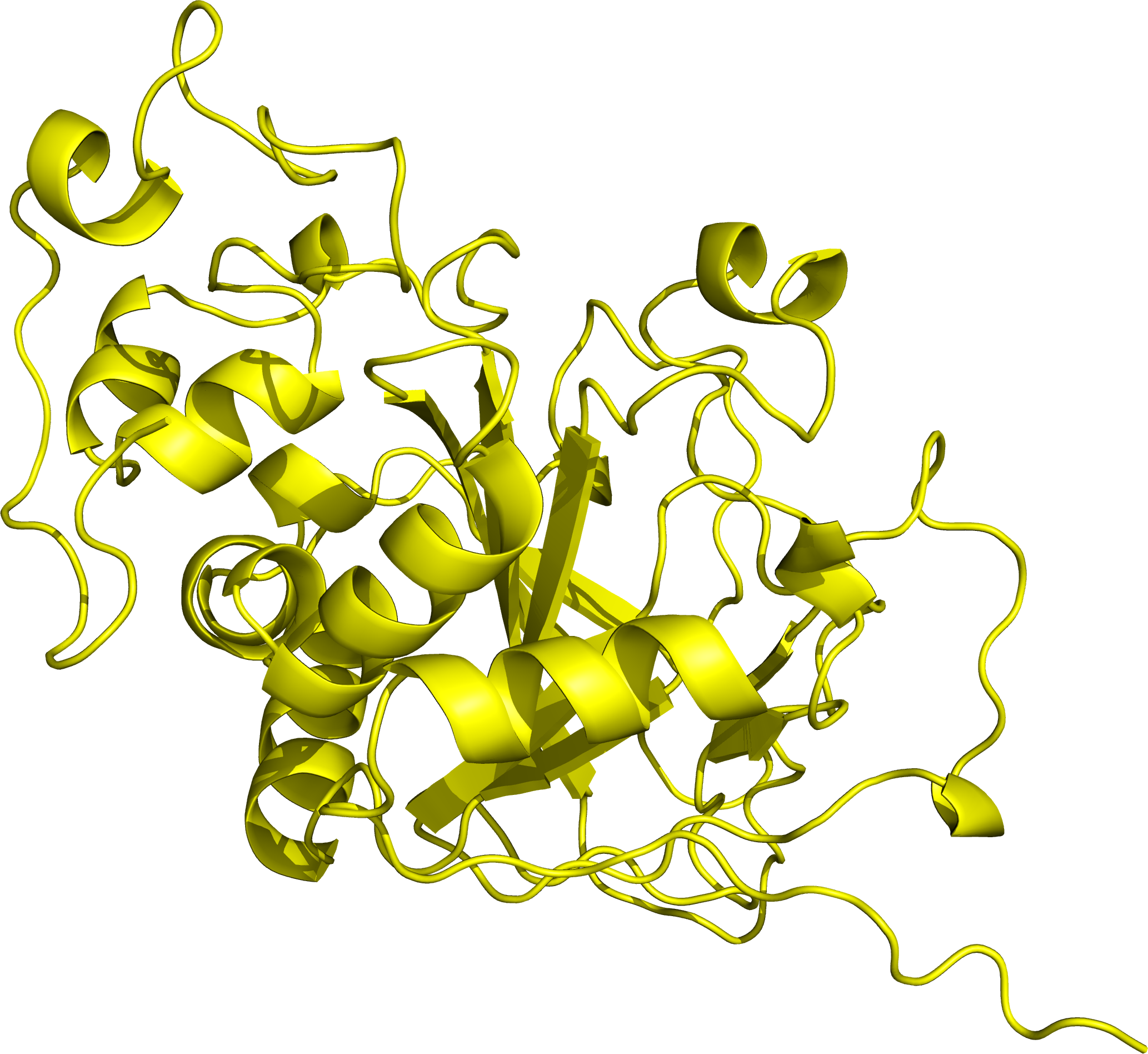}
    \caption{Structure of protein A0A0U5GJ41 (predicted using alphfold2).}
    \label{fig:j41}
  \end{figure}

  \begin{figure}[htbp!]
    \centering
    \includegraphics[width=0.75\linewidth]{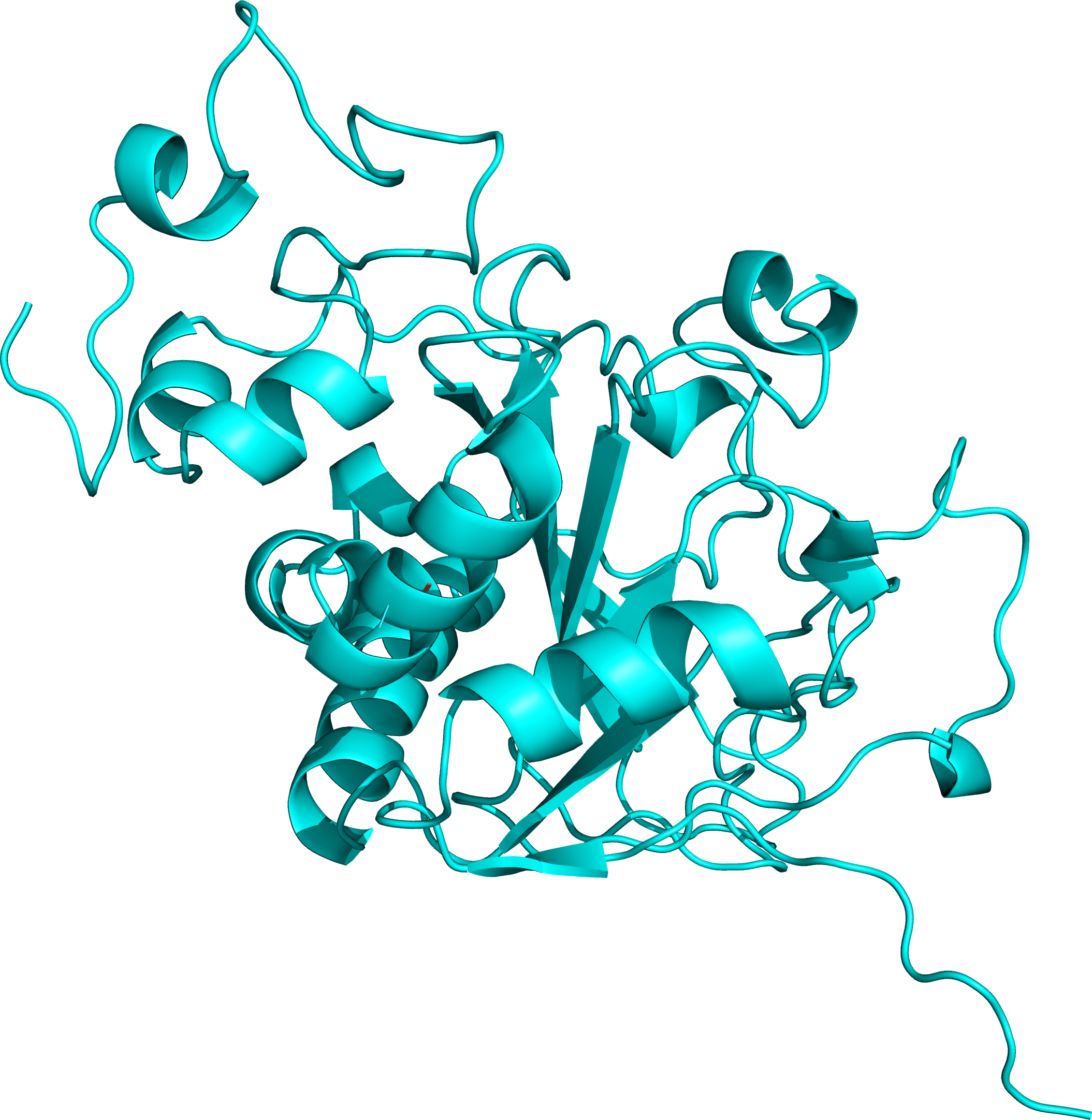}
    \caption{Structure of protein Q4WAW9 (predicted using alphfold2).}
    \label{fig:w9}
  \end{figure}

  \begin{figure}[htbp!]
    \centering
    \includegraphics[width=1.0\linewidth]{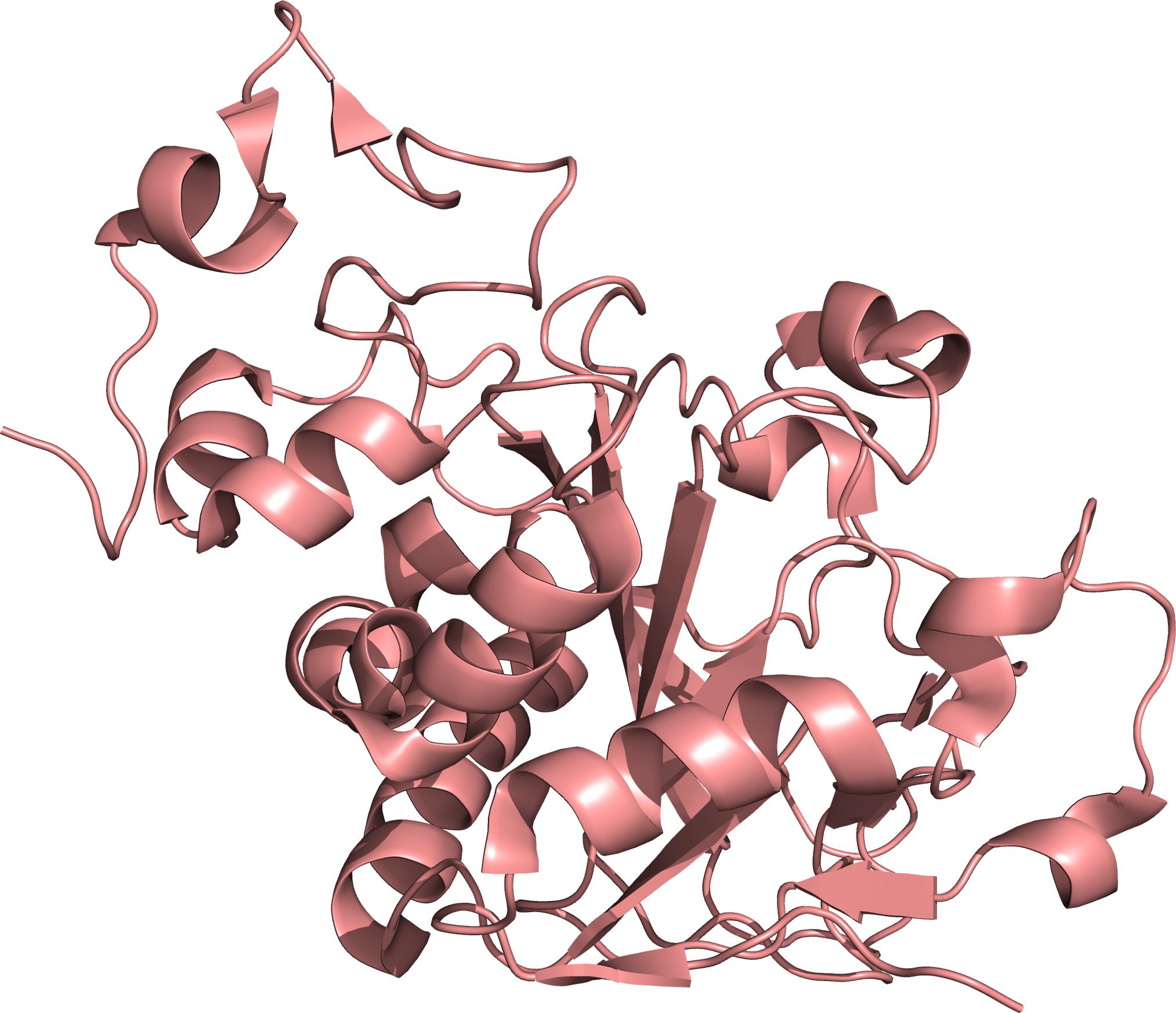}
    \caption{Structure for protein Q4WAW9 (alphfold2 predicted).}
    \label{fig:w9a}
  \end{figure}

  \begin{figure}[htbp!]
    \centering
    \includegraphics[width=1.0\linewidth]{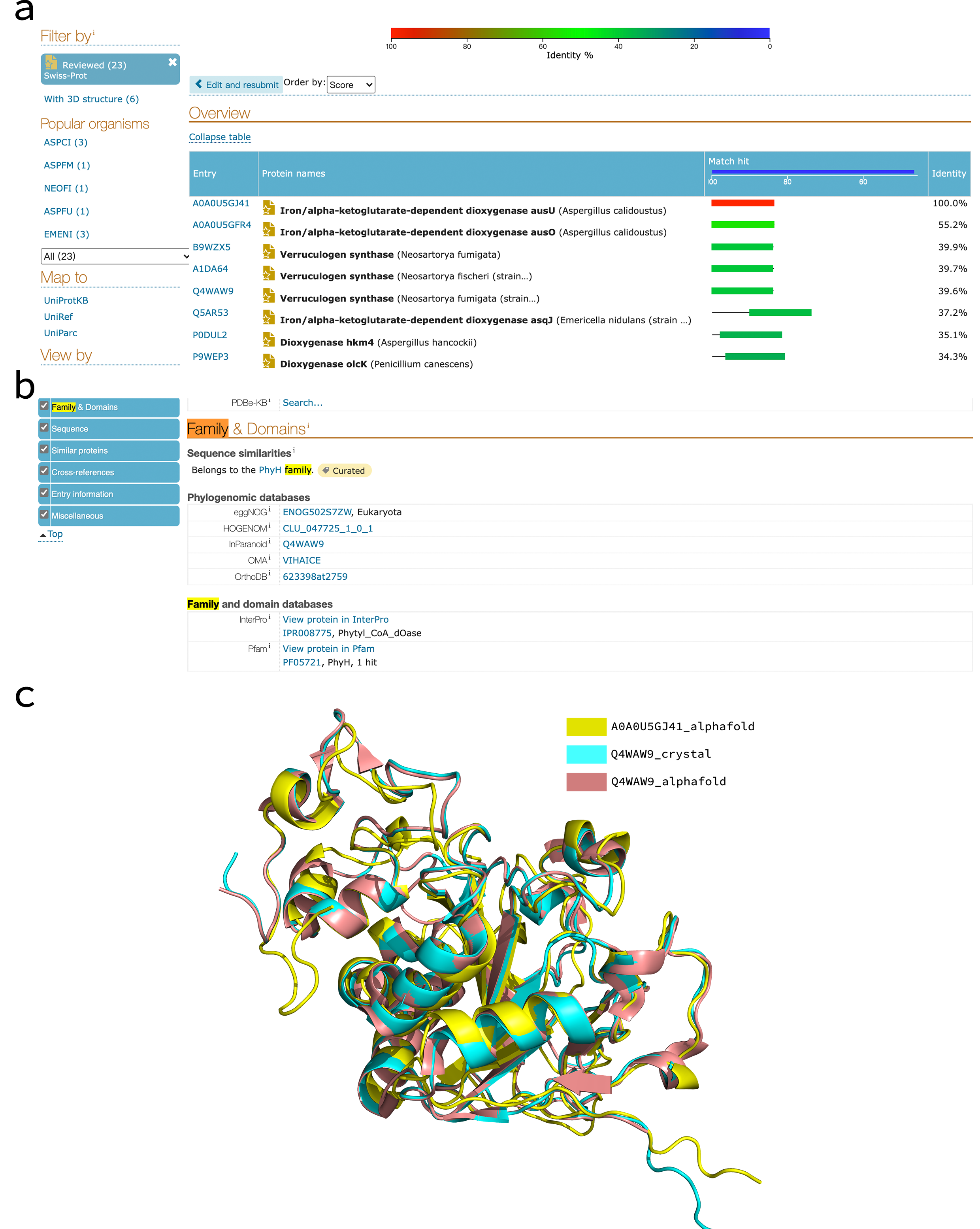}
    \caption{Comparison of sequence similarity and structural similarity.}
    \label{fig:case_merge}
  \end{figure}
\newpage

\section{SI Algorithm}

\begin{algorithm}[ht]
  \caption{Prepare benchmarking datasets}
  \label{alg:preparing_data_sets}
  \begin{algorithmic}[1]
    \State{download raw data from uniprot.org} \Comment{prepare\_task\_dataset.ipynb \# Step 3}
    \State \hspace{\algorithmicindent} {$train\_data$ $\gets$  $uniprot\_sprot-only2018\_02.tar.gz$ }
    \State \hspace{\algorithmicindent} {$test\_data$ $\gets$  $uniprot\_sprot-only2020\_06.tar.gz$ }
    \State{extract protein records from downloaded data} \Comment{prepare\_task\_dataset.ipynb \# Step 4}
    \State \hspace{\algorithmicindent} {extract protein $id$}
    \State \hspace{\algorithmicindent} {extract protein $name$}
    \State \hspace{\algorithmicindent} {extract protein $ec\_number$}
    \State \hspace{\algorithmicindent} {extract protein sequence as $seq$}
    \State \hspace{\algorithmicindent} {format $ec\_number$ and $seq$}
    \State \hspace{\algorithmicindent} {caculate protein arrtruibutes.} \Comment{exact\_ec\_from\_uniprot.py}
    \State{preprocessing protein records} \Comment{prepare\_task\_dataset.ipynb \# Step 6}
    \State \hspace{\algorithmicindent} {drop duplicates by $seq$}
    \State \hspace{\algorithmicindent} {remove changed $seq$ with same $id$}
    \State \hspace{\algorithmicindent} {format $ec\_number$ in standard four level like: -.-.-.-}
    \State \hspace{\algorithmicindent} {trim $ec\_number$ and $seq$ strings}
    \State{get esm embedding} \Comment{prepare\_task\_dataset.ipynb \# Step 6.6}
    \State{get unirep embeeding} \Comment{prepare\_task\_dataset.ipynb \# Step 6.7}
    \State{Construct task1 dataset $ds1$} \Comment{prepare\_task\_dataset.ipynb \# Step 7.1}
    \State{Construct task2 dataset $ds2$} \Comment{prepare\_task\_dataset.ipynb \# Step 7.2}
    \State{Construct task3 dataset $ds3$} \Comment{prepare\_task\_dataset.ipynb \# Step 7.3}
  \end{algorithmic}
\end{algorithm}

\begin{algorithm}[ht]
  \caption{EC Number Prediction}
  \label{alg:task3}
  \begin{algorithmic}[1]
  \State{load $trainset$ and $testset$ from $ds3$} \Comment{task3.ipynb \# Step 2}
  \State{load embedding features } \Comment{task3.ipynb \# Step 3}
  \State{conduct sequence alignment } \Comment{task3.ipynb \# Step 4}
  \State{transfer EC number to model labels} \Comment{task3.ipynb \# Step 5}
  \State{train EC prediction model} \Comment{task3.ipynb \# Step 6}
  \State {do EC prediction} \Comment{task3.ipynb \# Step 7} 
  \State{ \Return {prediction results}}
  \end{algorithmic}
\end{algorithm}

\begin{algorithm}[ht]
  \caption{Enzyme or Non-enzyme Prediction}
  \label{alg:task1}
  \begin{algorithmic}[1]
  \State{load $trainset$ and $testset$ from $ds1$} \Comment{task1.ipynb \# Step 2}
  \State{conduct sequence alignment  } \Comment{task1.ipynb \# Step 3}
  \State{embedding comparison } \Comment{task1.ipynb \# Step 4}
  \State \hspace{\algorithmicindent} {one-hot embedding} \Comment{task1.ipynb \# Step 4.1}
  \State \hspace{\algorithmicindent} {unirep embedding} \Comment{task1.ipynb \# Step 4.2}
  \State \hspace{\algorithmicindent} {esm layer 33 embedding} \Comment{task1.ipynb \# Step 4.3}
  \State \hspace{\algorithmicindent} {esm layer 32 embedding} \Comment{task1.ipynb \# Step 4.4}
  \State \hspace{\algorithmicindent} {esm layer 0 embedding} \Comment{task1.ipynb \# Step 4.5}
  \State{DMLF for enzyme or non-enzyme prediction  } \Comment{task1.ipynb \# Step 5}
  \State \hspace{\algorithmicindent} {learn model using KNN method on $train\_data$} 
  \State \hspace{\algorithmicindent} {predict enzyme or non-enzyme on $test\_data$ using learned model}
  \State \hspace{\algorithmicindent} {integrate KNN prediction with sequence alignment  prediction}
  \If{sequence alignment found homologous sequence}
    \State{use alignment results as prediction}
  \Else \\  \hspace{\algorithmicindent} {use KNN results as prediction}
  \EndIf
  \Return {prediction}
  \end{algorithmic}
\end{algorithm}

\begin{algorithm}[ht]
  \caption{Enzyme Catalytic Function Quantity Prediction}
  \label{alg:task2}
  \begin{algorithmic}[1]
  \State{load $trainset$ and $testset$ from $ds2$} \Comment{task2.ipynb \# Step 2}
  \State{load esm32 embedding features } \Comment{task2.ipynb \# Step 3}
  \State{conduct single or multi functions prediction benchmarking $sp$} \Comment{task2.ipynb \# Step 4.1}
  \State{conduct 2-8 functions prediction benchmarking $mp$} \Comment{task2.ipynb \# Step 4.2}
  \State {do sequences alignment} 
  \State {do function counts prediction} 
  \State {integrate and output results} \Comment{task2.ipynb \# Step 5.3}
  \If{sequence alignment found homologous sequence}
    \State{use alignment results as prediction}
  \ElsIf{$sp$ prediction is single functional}
    \State{prediction is $sp$ results}
  \Else \\  \hspace{\algorithmicindent} {use $mp$ results}
  \EndIf
  \Return {prediction results}
  \end{algorithmic}
\end{algorithm}

\newpage

\section{SI Appendix Tables}
\begin{table}[htbp]
  \centering
  \label{fig_si: benchmarking data}
  \caption{Benchmarking Data Description}
  \begin{tabular}{p{3cm}<{\centering}|c|c|c|c|c}
  \hline
  \rowcolor[HTML]{EFEFEF} &\multicolumn{2}{c|}{Snapshot}      & \multicolumn{3}{c}{Differ}  \\ \cline{2-6} 
  \rowcolor[HTML]{EFEFEF} 
  \multirow{-2}{*}{ITEM}  & February-2018 & June-2020 & difference & added & deleted        \\ \hline
  Records                 & 556,825       & 563,972   & 7147       & -     & -              \\ \hline
  Duplicate Removal       & 469,129       & 476,006   & 6877       & 8033  & 1156           \\ \hline
  Non-enzyme               & 246,562       & 247,319   & 757        & 4454  & 879            \\ \hline
  Enzyme                  & 222,565       & 228,687   & 6120       & 3579  & 277            \\ \hline
  Distinct EC             & 4854          & 5306      & 452        & 644   & 192            \\ \hline
  \end{tabular}
\end{table}

\begin{table}[htbp]
  \begin{adjustwidth}{-1.2in}{0in}
  \centering
  \label{fig_si: Task1_res}
  \caption{Task 1 Enzyme or None-enzyme Prediction Performance Commission}
  \begin{tabular}{c|c|c|c|c|c|r|r|r|r|r|r}
  \hline
  \rowcolor[HTML]{EFEFEF}    &  &   &    &    &    & \multicolumn{6}{c}{Confusion Matrix}           \\ \cline{7-12} 
  \rowcolor[HTML]{EFEFEF} 
  \multirow{-2}{*}{Baseline} & 
  \multirow{-2}{*}{ACC}  & 
  \multirow{-2}{*}{PPV}  & 
  \multirow{-2}{*}{NPV}  & 
  \multirow{-2}{*}{RC}   & 
  \multirow{-2}{*}{F1}   &TP      &FP      &FN      &TN     &UP    &UN                            \\ \hline
  ECPred        &0.7219  &0.8218  &0.9190  &0.8463  &0.8339 &3029  &657  &244  &277   &306  &1027 \\ \hline
  DeepEC        &0.6715  &0.9468  &0.6300  &0.2783  &0.4301 &996   &56   &2583 &4398  &0    &0    \\ \hline
  CatFam        &0.6502  &0.8050  &0.6214  &0.2836  &0.4194 &1015  &246  &2564 &4208  &0    &0    \\ \hline
  PRIAM${V2}$   &0.7410  &0.6486  &0.8967  &0.9137  &0.7586 &3270  &1772 &309  &2682  &0    &0    \\ \hline
  Ours          &0.9312  &0.9525  &0.9160  &0.8899  &0.9201 &3185  &159  &394  &4295  &0    &0    \\ \hline
  \end{tabular}
\end{adjustwidth}
\end{table}

\begin{table}[htpb]
  \centering
  \label{fig_si: Task2_res}
  \caption{Task 2 Multifunctional Enzyme Prediction Performance Commission}
  \begin{tabular}{p{2.8cm}<{\centering}|p{2cm}<{\centering}|p{2cm}<{\centering}|p{2cm}<{\centering}|p{2cm}<{\centering}}
  \hline
  \rowcolor[HTML]{EFEFEF} 
  basline     & accuracy & precision-macro & recall-macro & f1-macro \\ \hline
  ECPred      & 0.1444   & 0.8349          & 0.1673       & 0.0274   \\ \hline
  DeepEC      & 0.0852   & 0.8802          & 0.1360       & 0.0522   \\ \hline
  CatFam      & 0.1718   & 0.9172          & 0.1000       & 0.0293   \\ \hline
  PRIAM-V2    & 0.0462   & 0.0175          & 0.8564       & 0.0035   \\ \hline
  Ours        & 0.6454   & 0.5901          & 0.4444       & 0.3617   \\ \hline
  \end{tabular}
\end{table}

\begin{table}[htbp]
  \centering
  \label{fig_si: Task3_res}
  \caption{Task 3 EC Number Prediction Performance Commission}
  \begin{tabular}{c|c|c|c|c}
  \hline
  \rowcolor[HTML]{EFEFEF} 
  basline     & accuracy & precision-macro & recall-macro & f1-macro \\  \hline
  ECPred      & 0.0377   & 0.8042          & 0.2630       & 0.0908   \\ \hline
  DeepEC      & 0.0731   & 0.8121          & 0.3794       & 0.2376   \\ \hline
  CatFam      & 0.0705   & 0.8323          & 0.3507       & 0.2149   \\ \hline
  PRIAM-V2    & 0.0296   & 0.2080          & 0.7848       & 0.0220   \\ \hline
  Ours        & 0.8619   & 0.6900          & 0.6388       & 0.3676   \\ \hline
  \end{tabular}
\end{table}

\begin{table}[htbp]
  \begin{adjustwidth}{-1.8in}{0in}
  \centering
  \label{fig_si: embedding}
  \caption{Protein Sequences Embedding Performance Commission for Enzyme or Non-Enzyme}
  \begin{tabular}{c|c|c|c|c|c|c|l|l|l|l}
  \hline
  \rowcolor[HTML]{EFEFEF}    & & & & & & & \multicolumn{4}{c}{Confusion Matrix}  \\ \cline{8-11} 
  \rowcolor[HTML]{EFEFEF} 
  \multirow{-2}{*}{Method}   & 
  \multirow{-2}{*}{Baseline} & 
  \multirow{-2}{*}{ACC}      & 
  \multirow{-2}{*}{PPV}      & 
  \multirow{-2}{*}{NPV}      & 
  \multirow{-2}{*}{RC}       & 
  \multirow{-2}{*}{F1}       & TP   & FP     & FN     & TN \\ \hline
\multirow{5}{*}{Logistic Regression} & one-hot & 0.6473  & 0.5886  &0.7120   & 0.6924  & 0.6363 & 2478  & 1732 & 1101  & 2722  \\ \cline{2-11}
                                     & Unirep  & 0.8368  & 0.8593  & 0.8222  & 0.7578  & 0.8053 & 2712  & 444  & 867   & 4010  \\ \cline{2-11}
                                     & ESM0    & 0.7561  & 0.7209  & 0.7857  & 0.7385  & 0.7296 & 2643  & 1023 & 936   & 3431  \\ \cline{2-11}
                                     & ESM32   & 0.9066  & 0.9209  & 0.8964  & 0.8648  & 0.8919 & 3095  & 266  & 484   & 4188  \\ \cline{2-11}
                                     & ESM33   & 0.9032  & 0.9204  & 0.8909  & 0.8567  & 0.8874 & 3066  & 265  & 513   & 4189  \\ \hline \hline
\multirow{5}{*}{KNN}                 & one-hot & 0.6330  & 0.6686  & 0.6222  & 0.3495  & 0.4591 & 1251  & 620  & 2328  & 3834  \\ \cline{2-11}
                                     & Unirep  & 0.8486  & 0.8670  & 0.8363  & 0.7798  & 0.8211 & 2791  & 428  & 788   & 4026  \\ \cline{2-11}
                                     & ESM0    & 0.8246  & 0.7892  & 0.8556  & 0.8273  & 0.8078 & 2961  & 791  & 618   & 3663  \\ \cline{2-11}
                                     & ESM32   & 0.9294  & 0.9411  & 0.9208  & 0.8977  & 0.9189 & 3213  & 201  & 366   & 4253  \\ \cline{2-11}
                                     & ESM33   & 0.9273  & 0.9360  & 0.9208  & 0.8983  & 0.9167 & 3215  & 220  & 364   & 4234  \\ \hline \hline
\multirow{5}{*}{XGboost}             & one-hot & 0.7087  & 0.6851  & 0.7256  & 0.6407  & 0.6621 & 2293  & 1054 & 1286  & 3400  \\ \cline{2-11}
                                     & Unirep  & 0.8651  & 0.8885  & 0.8494  & 0.7972  & 0.8404 & 2853  & 358  & 726   & 4096  \\\cline{2-11}
                                     & ESM0    & 0.8282  & 0.8197  & 0.8346  & 0.7877  & 0.8034 & 2819  & 620  & 760   & 3834  \\\cline{2-11}
                                     & ESM32   & 0.9254  & 0.9540  & 0.9057  & 0.8748  & 0.9127 & 3131  & 151  & 448   & 4303  \\\cline{2-11}
                                     & ESM33   & 0.9157  & 0.9443  & 0.8962  & 0.8617  & 0.9011 & 3084  & 182  & 495   & 4272  \\\hline \hline
\multirow{5}{*}{Decision tree}       & one-hot & 0.6283  & 0.5889  & 0.6562  & 0.5488  & 0.5681 & 1964  & 1371 & 1615  & 3083  \\\cline{2-11}
                                     & Unirep  & 0.7966  & 0.7951  & 0.7976  & 0.7320  & 0.7623 & 2620  & 675  & 959   & 3779  \\\cline{2-11}
                                     & ESM0    & 0.7621  & 0.7437  & 0.7758  & 0.7111  & 0.7270 & 2545  & 877  & 1034  & 3577  \\\cline{2-11}
                                     & ESM32   & 0.8422  & 0.8550  & 0.8334  & 0.7776  & 0.8145 & 2783  & 472  & 796   & 3982  \\\cline{2-11}
                                     & ESM33   & 0.8311  & 0.8442  & 0.8223  & 0.7614  & 0.8006 & 2725  & 503  & 854   & 3951  \\\hline \hline
\multirow{5}{*}{Random forest}       & one-hot & 0.7162  & 0.6768  & 0.7493  & 0.6946  & 0.6856 & 2486  & 1187 & 1093  & 3267  \\\cline{2-11}
                                     & Unirep  & 0.8634  & 0.9151  & 0.8328  & 0.7645  & 0.8330 & 2736  & 254  & 843   & 4200  \\\cline{2-11}
                                     & ESM0    & 0.8539  & 0.8636  & 0.8470  & 0.7980  & 0.8295 & 2856  & 451  & 723   & 4003  \\\cline{2-11}
                                     & ESM32   & 0.9157  & 0.9657  & 0.8841  & 0.8407  & 0.8989 & 3009  & 107  & 570   & 4347  \\\cline{2-11}
                                     & ESM33   & 0.9161  & 0.9610  & 0.8871  & 0.8460  & 0.8999 & 3028  & 123  & 551   & 4331  \\\hline \hline
\multirow{5}{*}{GBDT}                & one-hot & 0.6775  & 0.6163  & 0.7461  & 0.7315  & 0.6690 & 2618  & 1630 & 961   & 2824  \\\cline{2-11}
                                     & Unirep  & 0.8332  & 0.8738  & 0.8091  & 0.7312  & 0.7962 & 2617  & 378  & 962   & 4076  \\\cline{2-11}
                                     & ESM0    & 0.8210  & 0.8100  & 0.8293  & 0.7815  & 0.7955 & 2797  & 656  & 782   & 3798  \\\cline{2-11}
                                     & ESM32   & 0.8720  & 0.9050  & 0.8507  & 0.7963  & 0.8472 & 2850  & 299  & 729   & 4155  \\\cline{2-11}
                                     & ESM33   & 0.8658  & 0.9017  & 0.8431  & 0.7843  & 0.8389 & 2807  & 306  & 772   & 4148  \\\hline 
  \end{tabular}
\end{adjustwidth}
\end{table}

\begin{table}[htbp]
  \begin{adjustwidth}{-0.8in}{0in}
  \centering
  \label{fig_si:embedding_poly}
  \caption{Protein Sequence Embedding Performance for Multifunctional Enzyme Prediction}
  \begin{tabular}{c|c|c|c|c|c}
  \hline
  \rowcolor[HTML]{EFEFEF}
  Baseline                             & Method  & ACC    & Precision-Macro & Recall-Macro & F1-Macro \\ \hline
  \multirow{5}{*}{Logistic regression} & One-hot & 0.9016 & 0.4485          & 0.2133       & 0.2206   \\ \cline{2-6} 
                                       & Unirep  & 0.9234 & 0.8462          & 0.1428       & 0.1372   \\ \cline{2-6} 
                                       & ESM0    & 0.9237 & 0.9891          & 0.1429       & 0.1372   \\ \cline{2-6} 
                                       & ESM32   & 0.9168 & 0.8205          & 0.3100       & 0.3719   \\ \cline{2-6}
                                       & ESM33   & 0.9210 & 0.7792          & 0.4365       & 0.4897   \\ \hline \hline
  \multirow{5}{*}{KNN}                 & One-hot & 0.9180 & 0.6498          & 0.3601       & 0.3511   \\ \cline{2-6} 
                                       & Unirep  & 0.9044 & 0.5790          & 0.1479       & 0.1474   \\ \cline{2-6} 
                                       & ESM0    & 0.9156 & 0.6195          & 0.4261       & 0.4672   \\ \cline{2-6} 
                                       & ESM32   & 0.9274 & 0.6317          & 0.5459       & 0.5773   \\ \cline{2-6} 
                                       & ESM33   & 0.9280 & 0.7974          & 0.5644       & 0.5994   \\ \hline \hline
  \multirow{5}{*}{XGboost}             & One-hot & 0.9252 & 0.8941          & 0.2374       & 0.2841   \\ \cline{2-6} 
                                       & Unirep  & 0.9192 & 0.8822          & 0.1480       & 0.1475   \\ \cline{2-6} 
                                       & ESM0    & 0.9258 & 0.8512          & 0.3878       & 0.4332   \\ \cline{2-6} 
                                       & ESM32   & 0.9389 & 0.9422          & 0.5101       & 0.5931   \\ \cline{2-6} 
                                       & ESM33   & 0.9380 & 0.9441          & 0.4626       & 0.5405   \\ \hline \hline
  \multirow{5}{*}{Decision tree}       & one-hot & 0.8593 & 0.3079          & 0.2185       & 0.2305   \\ \cline{2-6} 
                                       & Unirep  & 0.8647 & 0.5951          & 0.1430       & 0.1440   \\ \cline{2-6} 
                                       & ESM0    & 0.8786 & 0.5263          & 0.2531       & 0.2869   \\ \cline{2-6} 
                                       & ESM32   & 0.8874 & 0.3937          & 0.5412       & 0.3984   \\ \cline{2-6} 
                                       & ESM33   & 0.8814 & 0.3948          & 0.3862       & 0.2604   \\ \hline \hline
  \multirow{5}{*}{Random forest}       & One-hot & 0.9262 & 0.9419          & 0.2397       & 0.2887   \\ \cline{2-6} 
                                       & Unirep  & 0.9210 & 0.8462          & 0.1424       & 0.1370   \\ \cline{2-6} 
                                       & ESM0    & 0.9280 & 0.9421          & 0.3869       & 0.4317   \\ \cline{2-6} 
                                       & ESM32   & 0.9343 & 0.9394          & 0.4640       & 0.5398   \\ \cline{2-6} 
                                       & ESM33   & 0.9322 & 0.9271          & 0.4283       & 0.4997   \\ \hline \hline
  \multirow{5}{*}{GBDT}                & One-hot & 0.9125 & 0.1820          & 0.3125       & 0.1680   \\ \cline{2-6} 
                                       & Unirep  & 0.9228 & 0.8462          & 0.1427       & 0.1371   \\ \cline{2-6} 
                                       & ESM0    & 0.9240 & 0.5991          & 0.4407       & 0.3403   \\ \cline{2-6} 
                                       & ESM32   & 0.9271 & 0.6479          & 0.3135       & 0.3643   \\ \cline{2-6} 
                                       & ESM33   & 0.9231 & 0.6347          & 0.4828       & 0.3178   \\ \hline
  \end{tabular}
\end{adjustwidth}
  \end{table}

  \newpage

\nolinenumbers


\bibliographystyle{abbrv}